\documentclass[letterpaper, 10 pt, conference]{ieeeconf}  % Comment this line out if you need a4paper

\IEEEoverridecommandlockouts                              % This command is only needed if 
\usepackage[T1]{fontenc}
\usepackage{helvet}
\usepackage[pass]{geometry}
\usepackage{tcolorbox}
\usepackage{tikz}
\usepackage{booktabs}
\usepackage{colortbl}
\usepackage{xcolor}
\usepackage{graphicx}  % \resizebox 需要
\usepackage{amssymb}
\usepackage{amsmath}
\usepackage{cuted}
\usepackage{capt-of}
\usepackage{multirow}
\usepackage{makecell}
\usepackage[hidelinks]{hyperref}

\begin{document}
\pagestyle{empty}
\newgeometry{left=0.75in,right=0.55in,top=0.45in,bottom=0.45in}
\onecolumn
\thispagestyle{empty}
% Full-width first page. Institution logo filenames match affiliation superscripts.
% Fill in these four URLs when available. Empty URLs leave plain-text labels.
\newcommand*{\ProjectURL}{https://tacclaw-infinity.github.io/}
\newcommand*{\GitHubURL}{https://github.com/tacclaw-infinity/TacClaw-Infinity}
\newcommand*{\ModelScopeURL}{https://modelscope.cn/datasets/daimonrobotics/Daimon-Infinity}
\newcommand*{\HuggingFaceURL}{https://huggingface.co/datasets/daimonrobotics/Daimon-Infinity}
\newcommand*{\emptyresourceurl}{}
\newcommand{\paperresourcelink}[2]{%
  \ifx#1\emptyresourceurl #2\else\href{#1}{#2}\fi}
% Envelope scales with the surrounding text, including author superscripts.
\newcommand{\correspondingmark}{%
  \tikz[x=1em,y=1em,baseline=-0.06em,line width=0.045em,line join=round]{%
    \draw (0,0) rectangle (0.8,0.52);
    \draw (0,0.52) -- (0.4,0.20) -- (0.8,0.52);}}
% Missing logo files leave an empty slot; add NN.png to fill a slot.
% White paper backgrounds in legacy raster artwork are composited away using
% PDF multiply blending, preserving the original emblem pixels and fine lettering.
% Optional graphicx trim clips wordmarks or excess margins at placement time.
\newcommand{\institutionlogo}[2][]{%
  \begin{minipage}[c]{0.057\linewidth}\centering
  \parbox[c][25pt][c]{\linewidth}{\centering
    \IfFileExists{figures/institution-logos/#2.png}{%
      \tikz[baseline=(mark.base)]{\node[inner sep=0pt,outer sep=0pt,blend mode=multiply] (mark) {%
        \includegraphics[#1,width=0.96\linewidth,height=25pt,keepaspectratio]{figures/institution-logos/#2.png}};}%
    }{\rule{0pt}{25pt}}}%
  \end{minipage}%
}
\definecolor{frontmatterbg}{HTML}{F0F4F6}
\begin{tcolorbox}[colback=frontmatterbg,colframe=frontmatterbg,
  boxrule=0pt,arc=9pt,left=20pt,right=20pt,top=18pt,bottom=14pt,
  before skip=0pt,after skip=0pt]
{\sffamily\bfseries\fontsize{16.5}{19.5}\selectfont\raggedright
TacClaw-Infinity: A Large-Scale Vision-Tactile-\allowbreak{}Language-Action Dataset for Dexterous Manipulation\par}
\vspace{10pt}
{\sffamily\bfseries\fontsize{10}{12}\selectfont\raggedright
\mbox{Fucai Zhu\textsuperscript{1,3,*}}, \mbox{Xidong Zhang\textsuperscript{1,3,*}}, \mbox{Yichi Zhang\textsuperscript{1,*}}, \mbox{Yipan Wang\textsuperscript{5}}, \mbox{Fubo Wang\textsuperscript{3}},  \mbox{Zibin Wu\textsuperscript{6}}, \mbox{Kaiqi Li\textsuperscript{7}}, \mbox{Jianhao Jiao\textsuperscript{8}}, \mbox{Shihan Lu\textsuperscript{9}}, \mbox{Di Hu\textsuperscript{10}}, \mbox{Mengdi Xu\textsuperscript{11}},  \mbox{Siyu Zhu\textsuperscript{12}}, \mbox{Xiaoxiao Long\textsuperscript{2}}, \mbox{Yao Yao\textsuperscript{2}}, \mbox{Xun Cao\textsuperscript{2}}, \mbox{Zhuo Xu\textsuperscript{13}},  \mbox{Koichi Hashimoto\textsuperscript{4}}, \mbox{Qifeng Chen\textsuperscript{14}}, \mbox{Zhe Li\textsuperscript{7}}, \mbox{Jianfei Yang\textsuperscript{7}}, \mbox{Haofei Lu\textsuperscript{15}},  \mbox{Danica Kragic\textsuperscript{15}}, \mbox{Shuicheng Yan\textsuperscript{16}}, \mbox{Jianghua Duan\textsuperscript{1}}, \mbox{Michael Yu Wang\textsuperscript{1,3}}, \mbox{Weihao Yuan\textsuperscript{1,2,\correspondingmark}}\par}
\vspace{9pt}
{\normalfont\fontsize{8}{9.7}\selectfont\raggedright
\textsuperscript{1}Daimon Robotics \quad \textsuperscript{2}Nanjing University \quad \textsuperscript{3}Great Bay University\\
\textsuperscript{4}Tohoku University \quad \textsuperscript{5}Southern University of Science and Technology\\
\textsuperscript{6}The Hong Kong University of Science and Technology (Guangzhou)\\
\textsuperscript{7}Nanyang Technological University \quad \textsuperscript{8}University College London \quad \textsuperscript{9}Northwest University\\
\textsuperscript{10}Renmin University of China \quad \textsuperscript{11}Tsinghua University \quad \textsuperscript{12}Fudan University\\
\textsuperscript{13}DeepMind \quad \textsuperscript{14}The Hong Kong University of Science and Technology\\
\textsuperscript{15}Royal Institute of Technology \quad \textsuperscript{16}National University of Singapore\par}
\vspace{3pt}
{\normalfont\fontsize{7.3}{9}\selectfont
\textsuperscript{*} Equal Contributions,\quad \correspondingmark\ Corresponding Author\par}
\vspace{10pt}
{\normalfont\fontsize{10}{12}\selectfont
\renewenvironment{abstract}{\par\noindent\ignorespaces}{\par}
\setlength{\parindent}{0pt}\setlength{\parskip}{0pt}
\begin{abstract}
% 1 Background
% 1.1 大规模 robot learning 需要多样化 human demonstrations
% 1.2 tactile manipulation dataset 仍然难以规模化;且大规模异构数据往往存在质量不一致、缺乏结构化监督的问题
Large-scale robot learning increasingly relies on diverse human demonstrations, yet tactile manipulation data remain difficult to scale, and existing multimodal datasets often lack consistent quality control and structured supervision.
% 2 Dataset overview
% 2.1 直接提出 TacClaw-Infinity，把 dataset 放到摘要最前面
% 2.2 TacClaw 只作为数据规模化采集的 enabling interface 简短介绍
% 2.3 强调 synchronized vision-tactile-language-action，以及不依赖目标机器人采集
We present \textbf{TacClaw-Infinity}, a large-scale structured Vision-Tactile-Language-Action (VTLA) dataset for contact-rich manipulation, collected with \textbf{TacClaw}, a tactile-enabled UMI-style handheld interface that records synchronized visual observations, fingertip tactile signals, and end-effector trajectories without requiring a target robot during data collection.
% 3 Scale & Diversity
% 3.1 两种互补采集模式：centralized protocol-driven + portable distributed collection
% 3.2 前者保证高吞吐、可重复性和任务覆盖，后者提升场景、机构、应用分布的多样性
% 3.3 distributed collection 覆盖 homes / outdoors / labs / industrial workplaces，以及合作机构自定义任务
% 3.4 这里给核心规模数字
TacClaw-Infinity is built through a distributed worldwide collection across homes, laboratories, industrial workplaces, retail stores, and outdoor environments  including tasks contributed by academic and industrial collaborators.
Together, these efforts yield over \textbf{10,000 hours}, \textbf{1 million episodes}, and \textbf{1 billion frames} spanning more than \textbf{1,000 task types}.
% , with 400+ hour data collected outside centralized facilities.
% 4 Data quality & Structure
% 4.1 curation: trajectory reconstruction + multimodal QC + embodiment-aware feasibility filtering
% 4.2 annotation: 结构化标注包括 contact-centric segmentation, subtask description, and spatial grounding
We develop a scalable pipeline for trajectory reconstruction, multimodal quality control, embodiment-aware filtering, and structured annotation, covering contact-centric segmentation, subtask description, and spatial grounding.
% 5 Experiment
% 5.1 数据层面：验证硬件，curation和annotation的有效性
% 5.1 模型层面：仅用 TacClaw 数据训练/后训练，跨多个 embodiment 直接部署
We validate TacClaw-Infinity at both the data and model levels: At the data level, we assess the collection interface, the action-space compatibility of the curated demonstrations, and the structured annotations, while for the model level, we demonstrate cross-embodiment transfer across three heterogeneous robots without robot-specific demonstrations.
% \weihao{For pipeline and exp, similar or even richer than intro. intro needs richen and here needs simplify} 
\end{abstract}

% To make this heterogeneous corpus reliable and learning-ready, we develop a scalable processing pipeline for trajectory reconstruction, multimodal quality control, embodiment-aware feasibility filtering across multiple robot embodiments, and automatic structured annotation. The resulting temporal, semantic, and spatial supervision includes contact-centric subtask segmentation, segment-level language descriptions, and contact-anchored target-object localization and mask tracking.

% We evaluate TacClaw-Infinity using two policy families across three heterogeneous robot embodiments. Policies trained solely on TacClaw demonstrations transfer without embodiment-specific robot demonstrations, and controlled comparisons isolate the contribution of tactile observations and structured annotations.
}
\vspace{8pt}
{\sffamily\bfseries\fontsize{9}{11}\selectfont\centering\color{blue!45!black}
\paperresourcelink{\ProjectURL}{Project}\quad\textbullet\quad
\paperresourcelink{\GitHubURL}{GitHub}\quad\textbullet\quad
\paperresourcelink{\ModelScopeURL}{ModelScope}\quad\textbullet\quad
\paperresourcelink{\HuggingFaceURL}{Hugging Face}\par}
\vspace{6pt}
{\centering
\institutionlogo{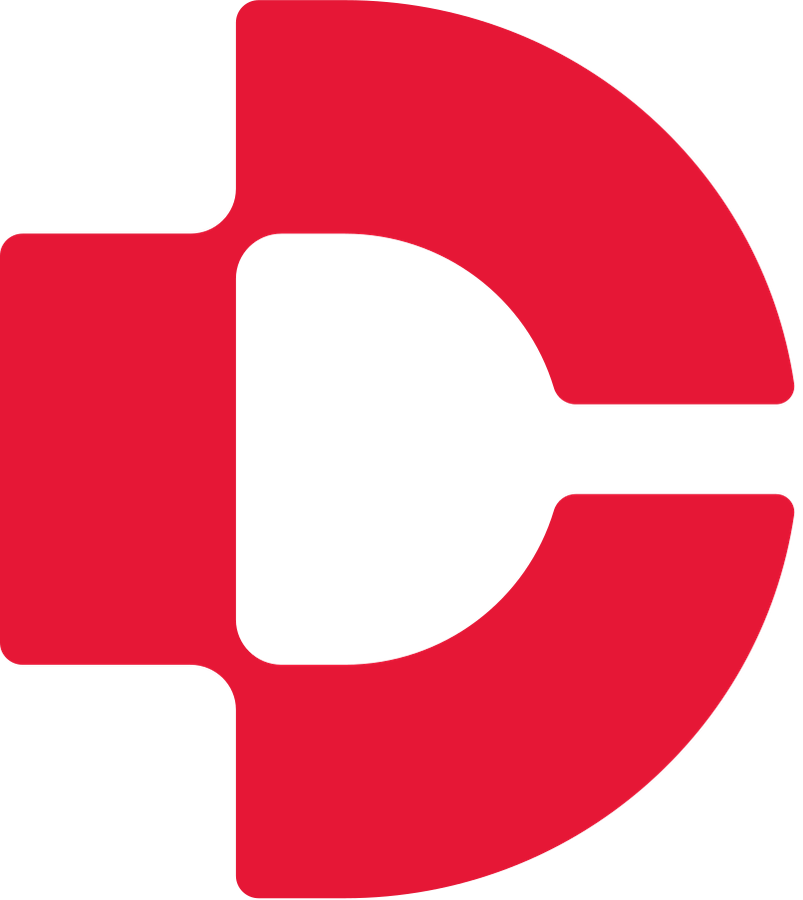}\hfill\institutionlogo{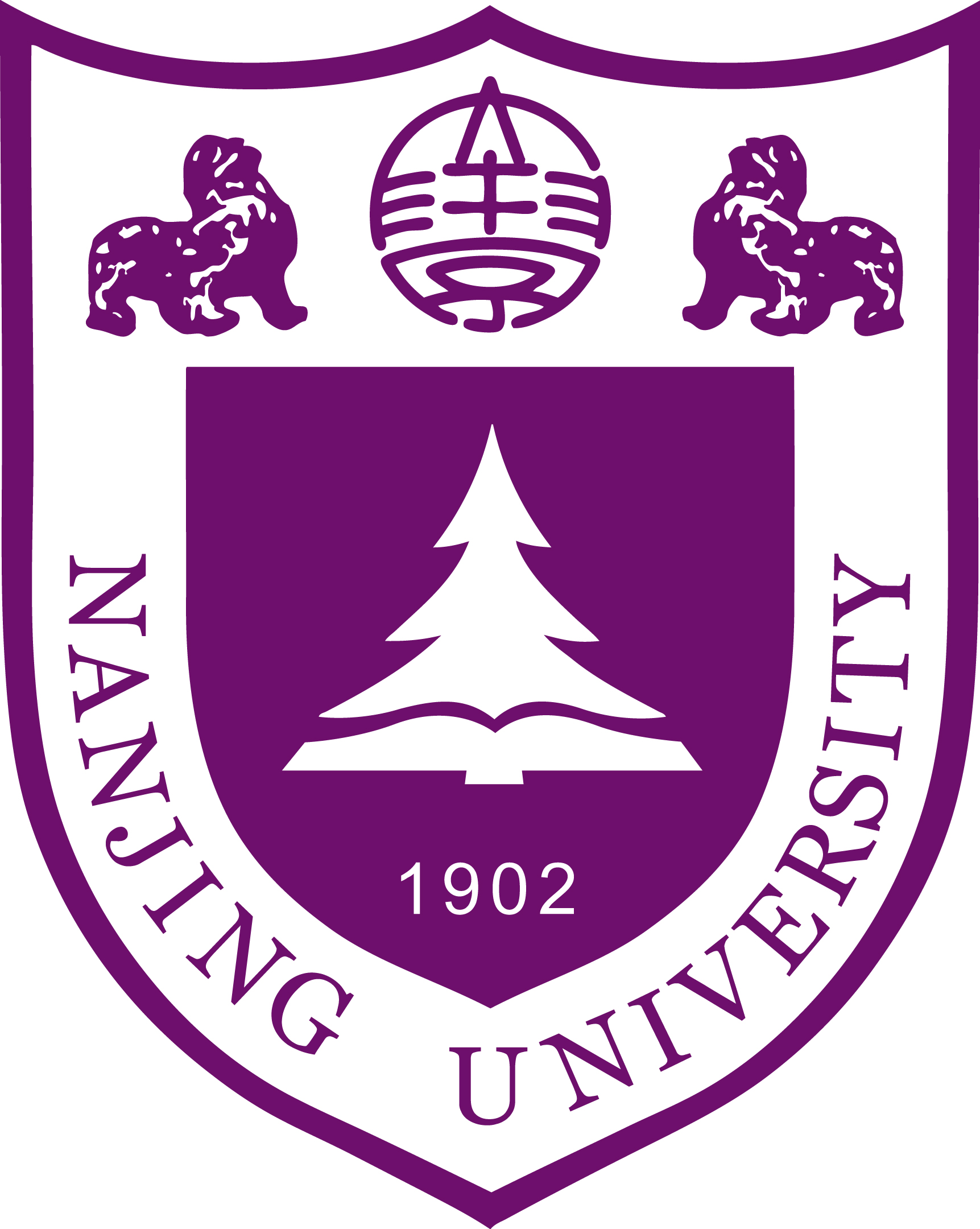}\hfill\institutionlogo{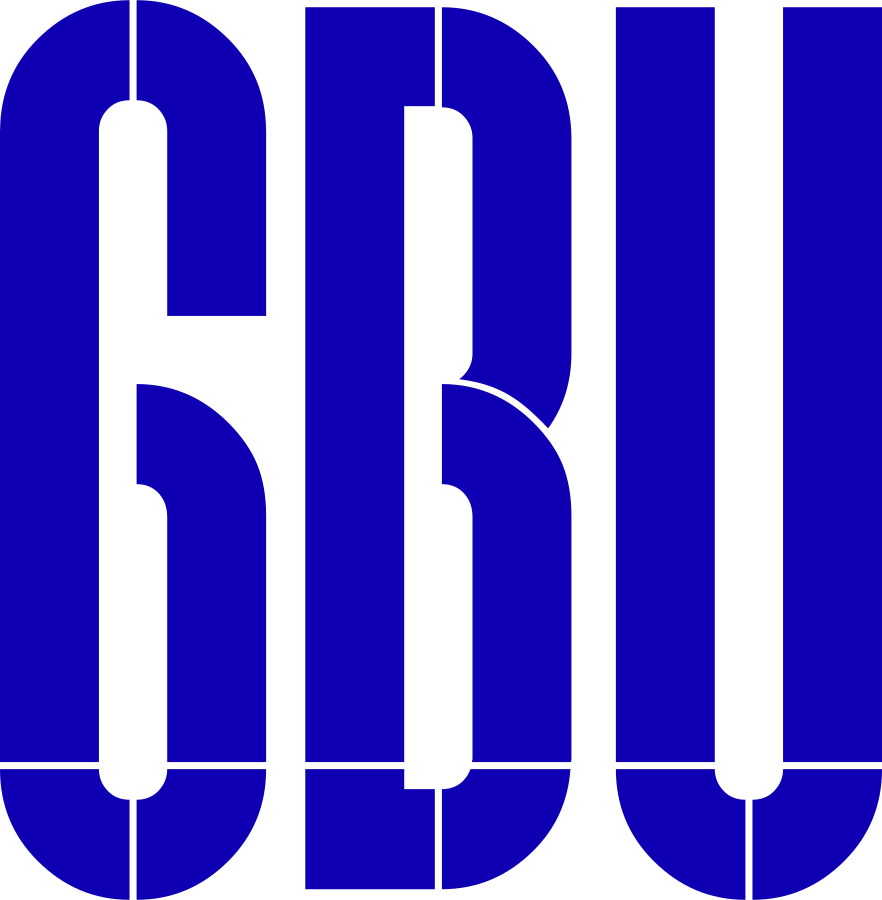}\hfill\institutionlogo{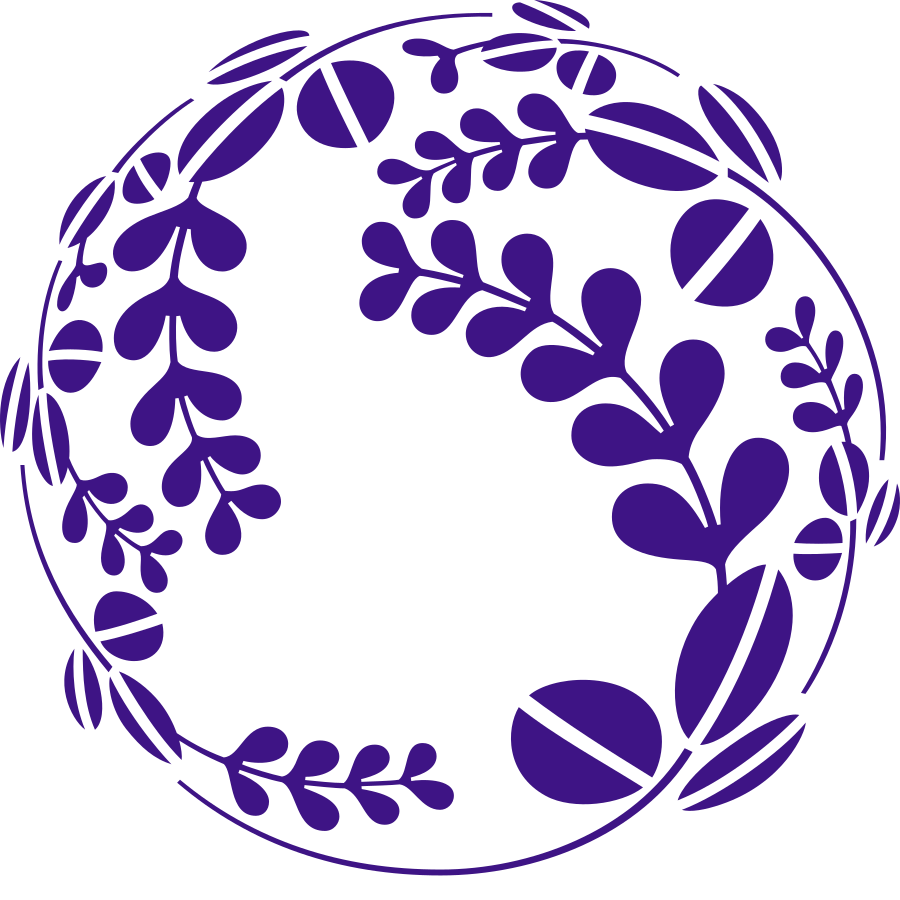}\hfill\institutionlogo{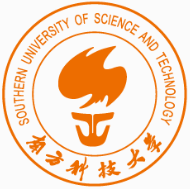}\hfill\institutionlogo{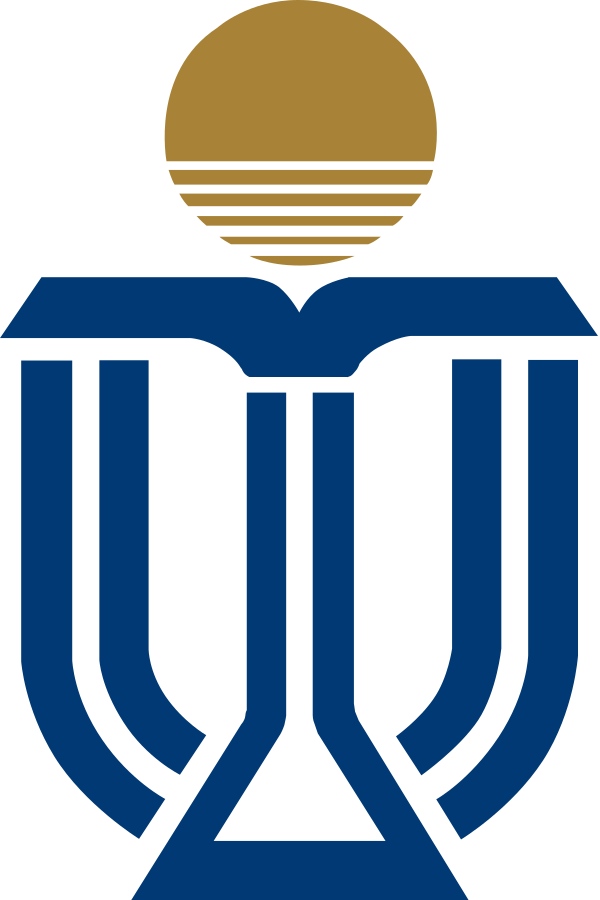}\hfill\institutionlogo[trim=0 0 519.84bp 0,clip]{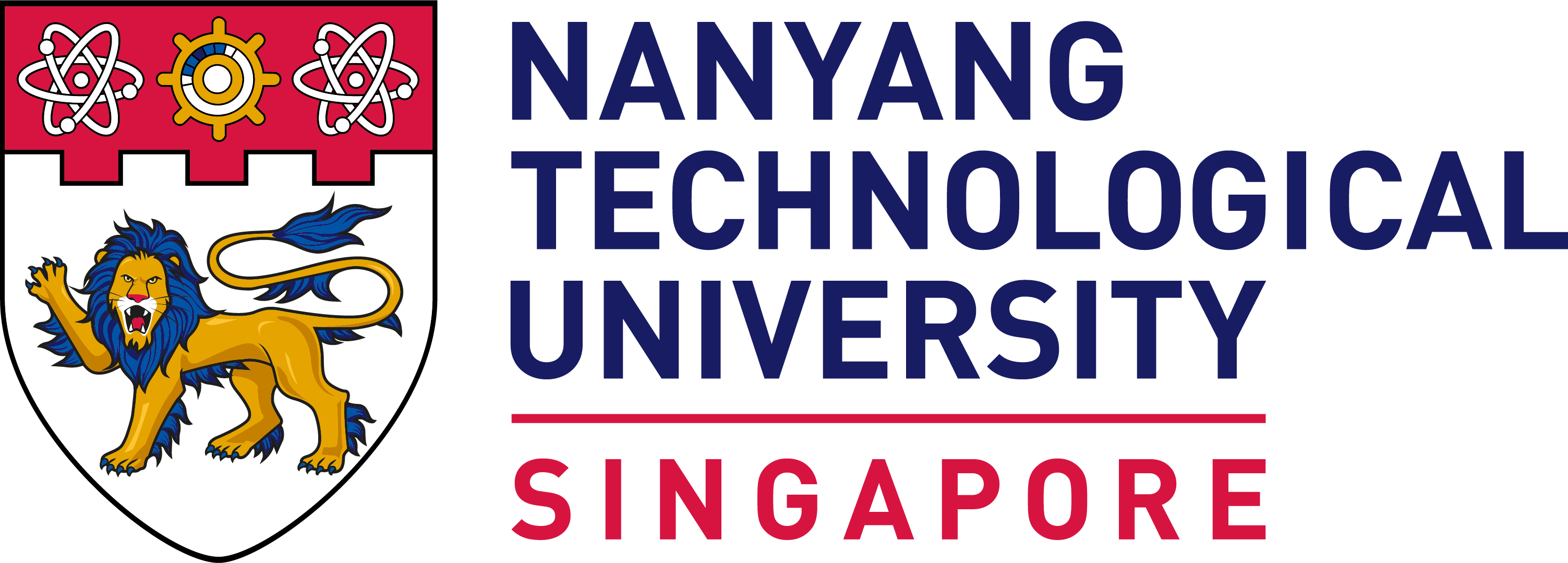}\hfill\institutionlogo[trim=23bp 18bp 27bp 17bp,clip]{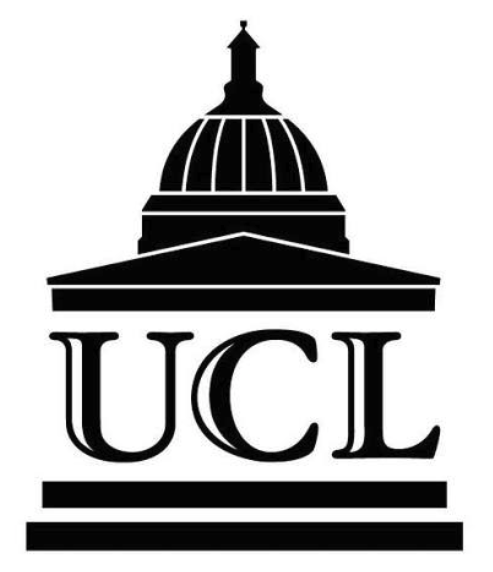}\hfill\institutionlogo{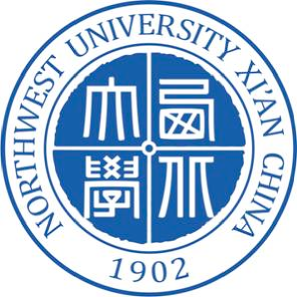}\hfill\institutionlogo{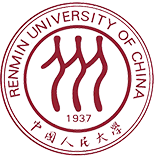}\hfill\institutionlogo{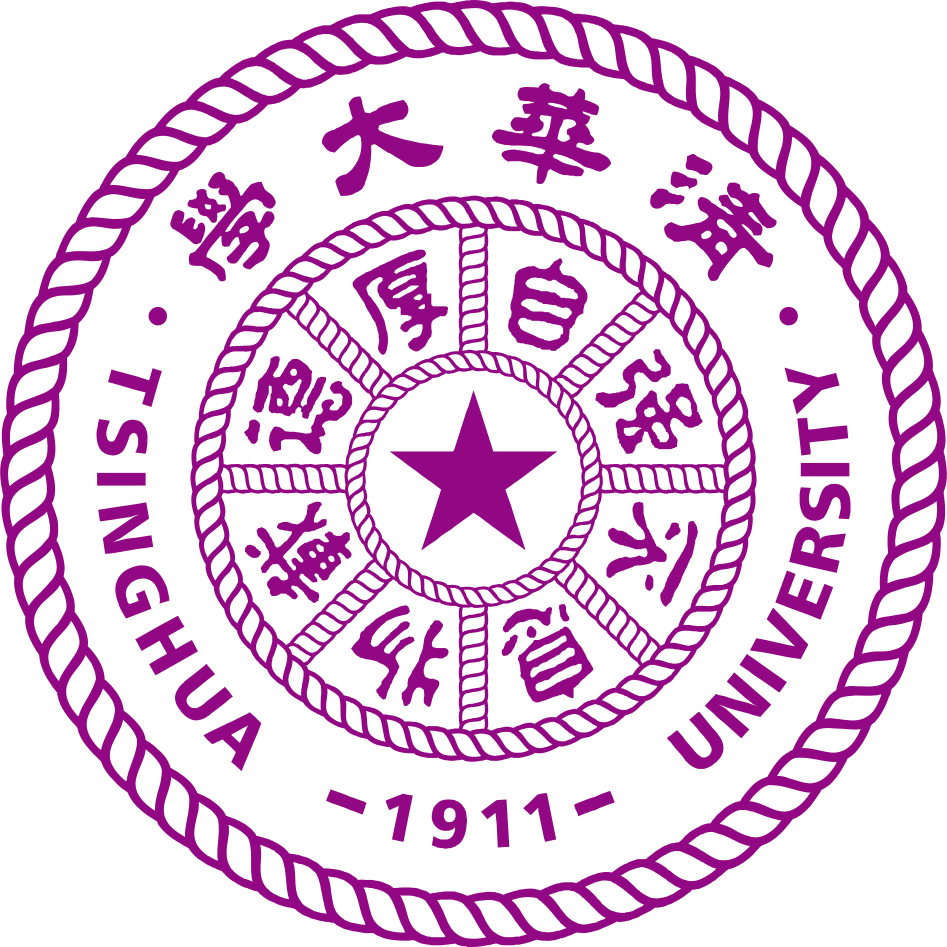}\hfill\institutionlogo{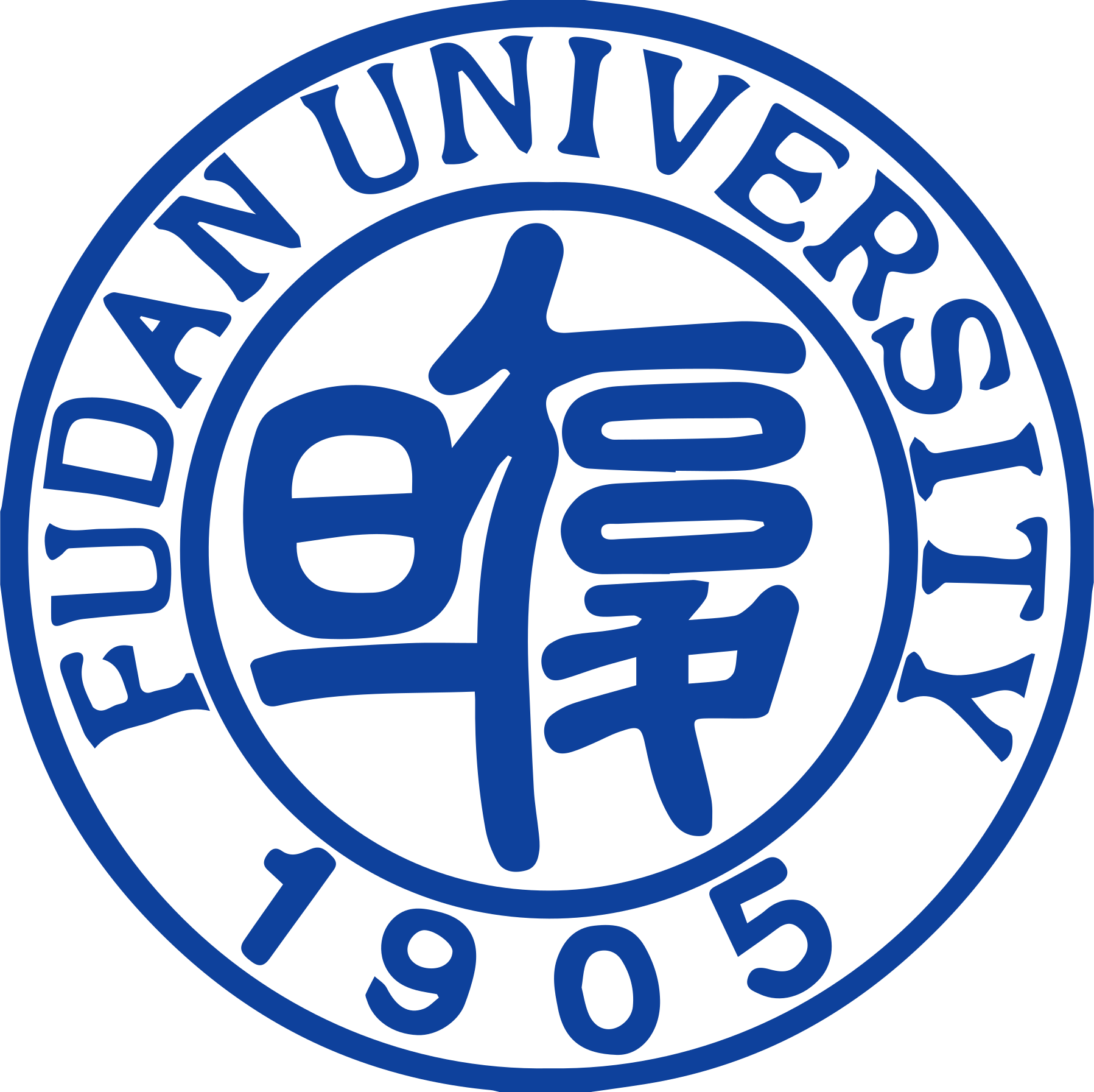}\hfill\institutionlogo{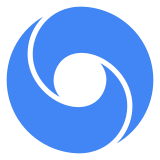}\hfill\institutionlogo{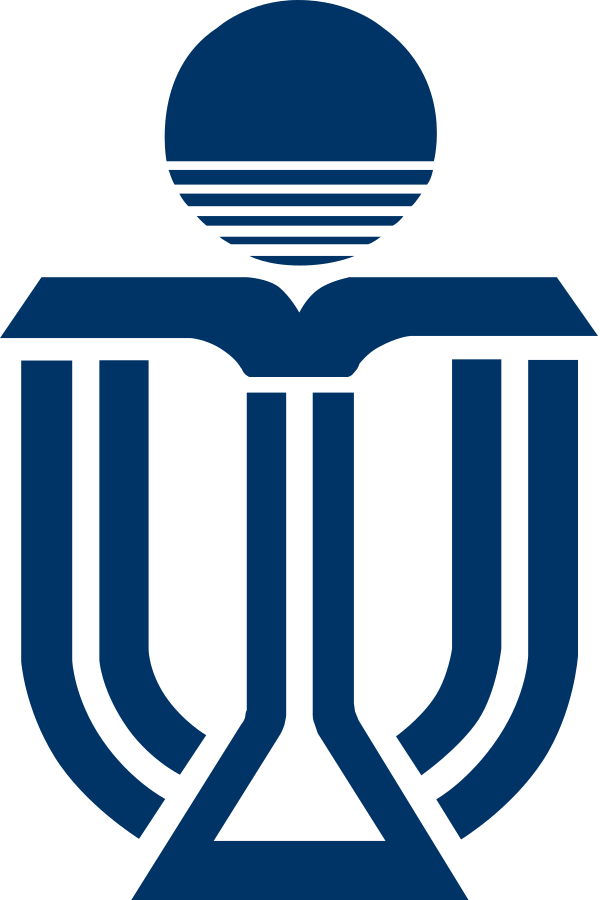}\hfill\institutionlogo{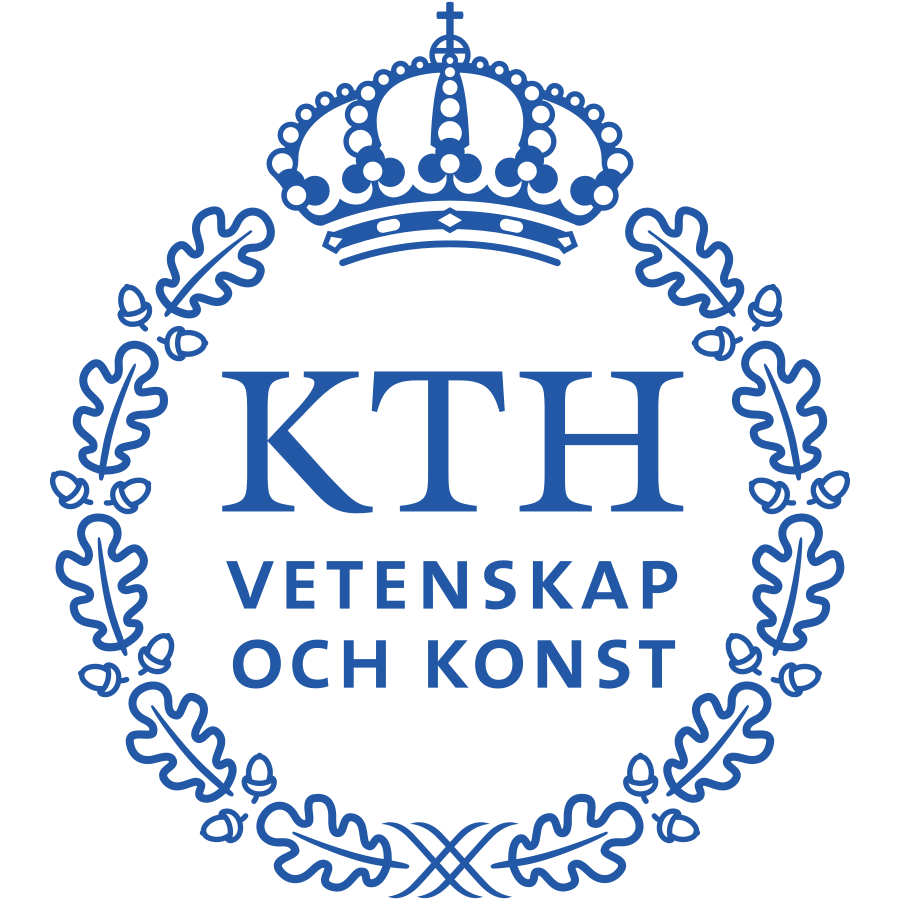}\hfill\institutionlogo[trim=74bp 19bp 72bp 33bp,clip]{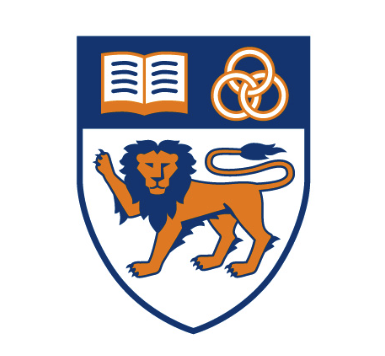}\par}
\end{tcolorbox}
\par\vspace{12.85pt}\nointerlineskip
\noindent\begin{minipage}{\textwidth}
\centering
\includegraphics[width=\textwidth]{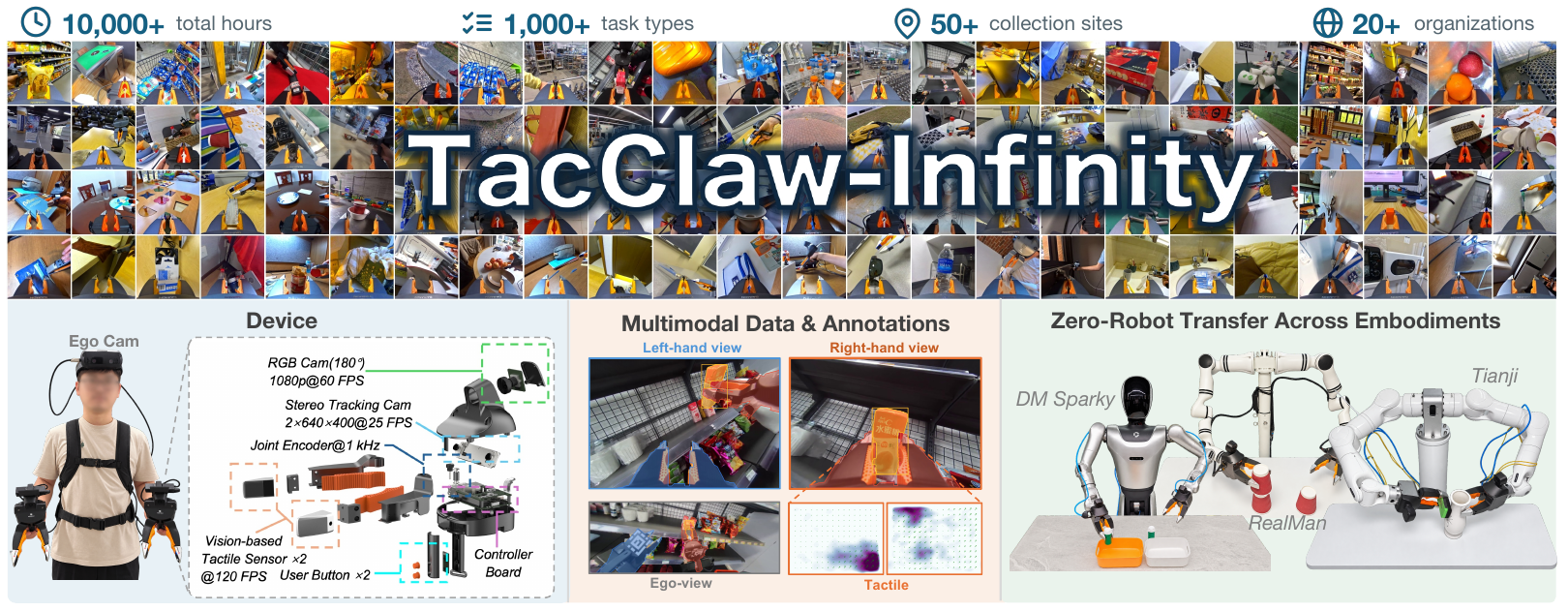}
\par\vspace{8pt}
{\normalfont\fontsize{7.8}{9.2}\selectfont\raggedright
\refstepcounter{figure}\label{fig:teaser}%
\textbf{Fig.~\thefigure. Overview of TacClaw-Infinity.}
A tactile-enabled UMI-style interface supports distributed large-scale collection of synchronized visual, tactile, language, and action data across diverse tasks and environments. The resulting dataset enables zero-robot policy transfer across different heterogeneous robot embodiments.\par}
\end{minipage}

\restoregeometry
\twocolumn

\section{Introduction}

% 1 Background
% 1.1 大规模 robot learning 需要多样化 human demonstrations
% 1.2 tactile manipulation dataset 仍然难以规模化;且大规模异构数据往往存在质量不一致、缺乏结构化监督的问题
Large-scale robot learning increasingly relies on diverse human demonstrations to support generalizable manipulation across tasks, objects, and environments. While recent vision-language-action (VLA) models benefit from large-scale multimodal data, contact-rich manipulation additionally requires tactile information that is difficult to infer from vision alone, especially under occlusion, sustained contact, or fine-grained interaction. However, tactile manipulation data remain difficult to scale because collection is often tied to robot-specific sensor integration, calibration, and teleoperation. Moreover, large-scale multimodal demonstrations can vary substantially in quality and typically provide limited structured supervision for contact, task progress, and interaction targets.

% 2 Dataset overview
% 2.1 直接提出 TacClaw-Infinity，把 dataset 放到摘要最前面
% 2.2 TacClaw 只作为数据规模化采集的 enabling interface 简短介绍
% 2.3 强调 synchronized vision-tactile-language-action，以及不依赖目标机器人采集
To address these challenges, we introduce \textbf{TacClaw-Infinity},
a large-scale structured Vision-Tactile-Language-Action (VTLA) dataset
for contact-rich manipulation, as illustrated in
Fig.~\ref{fig:teaser}. The dataset is collected with \textbf{TacClaw}, a tactile-enabled UMI-style handheld interface that synchronously records visual observations, fingertip tactile signals, and end-effector trajectories without requiring a target robot during data collection. This robot-independent design enables scalable acquisition across different operators and environments while preserving rich physical interaction signals.

% 3 Scale & Diversity
% 3.1 两种互补采集模式：centralized protocol-driven + portable distributed collection
% 3.2 前者保证高吞吐、可重复性和任务覆盖，后者提升场景、机构、应用分布的多样性
% 3.3 distributed collection 覆盖 homes / outdoors / labs / industrial workplaces，以及合作机构自定义任务
% 3.4 这里给核心规模数字  over 0,000 hours, 1 million episodes, and 1 billion frames spanning more than 6,000 task type
To achieve both scale and diversity, TacClaw-Infinity is collected at more than 50 physical locations spanning homes, laboratories, industrial workplaces, retail stores, outdoor environments, and other everyday settings around the world. Leveraging the portability of TacClaw, we combine protocol-driven recordings of common skills from a shared task catalog with open-ended tasks chosen by collectors.
Worldwide academic and industrial collaborating organizations further contribute tasks according to their own research and application interests. Together, these efforts yield data of over \textbf{10,000 hours}, \textbf{1 million episodes}, and \textbf{1 billion frames} spanning more than \textbf{1,000 task types}. 
% To the best of our knowledge, TacClaw-Infinity is the largest visuotactile manipulation dataset reported to date, in terms of both demonstration scale and task diversity. 
% spanning more than \textbf{1,000 task types}, with over 400 hour data collected outside centralized facilities.\weihao{what does this mean?}
% \weihao{To the best of our knowledge, this is the largest xxxxxxx}

% 4 Data quality & Structure
% 4.1 pipeline 不作为主角，只说明解决 large-scale heterogeneous data 的质量与可用性问题
% 4.2 curation: trajectory reconstruction + multimodal QC + embodiment-aware feasibility filtering
% 4.3 annotation: temporal + semantic + spatial structured supervision
% 4.4 结构化标注包括 contact-centric segmentation、segment language、target object tracking
To make the heterogeneous corpus reliable and learning-ready, we develop a scalable processing pipeline covering trajectory reconstruction, multimodal quality control, and embodiment-aware feasibility filtering based on the reachability and manipulability of multiple robot embodiments. We further provide automatic temporal, semantic, and spatial annotations, including contact-centric subtask segmentation, segment-level language descriptions, and contact-anchored target-object localization and mask tracking. These annotations transform raw multimodal streams into structured supervision for downstream policy learning.

% 5 Experiment
% 5.1 数据层面：验证硬件，curation和annotation的有效性
% 5.1 模型层面：仅用 TacClaw 数据训练/后训练，跨多个 embodiment 直接部署
We validate TacClaw-Infinity from both data and model perspectives. At the data level, we evaluate the haptic fidelity of the TacClaw interface, characterize the cross-embodiment action-space compatibility of the collected demonstrations, and assess the quality of the structured annotations. At the model level, we evaluate two representative policy families across three heterogeneous robot embodiments and demonstrate that policies trained solely on TacClaw-Infinity can transfer without embodiment-specific demonstrations. Controlled comparisons further isolate the contributions of tactile observations to downstream policy learning.

Our main contributions are summarized as follows:
\begin{itemize}
\item We introduce \textbf{TacClaw-Infinity}, a large-scale structured VTLA dataset spanning more than \textbf{1,000 task types}, collected in a distributed manner across \textbf{50+ locations} and \textbf{20+ organizations} around the world.
\item We develop a scalable robot-independent collection and processing pipeline with trajectory reconstruction, embodiment-aware feasibility assessment, and automatic temporal, semantic, and spatial annotations.
We further evaluate the collection interface, action-space compatibility, and annotation quality.
\item We validate the dataset across three robot embodiments and two policy families, demonstrating both cross-embodiment transferability and the effectiveness of tactile sensing.
\end{itemize}

% We evaluate TacClaw-Infinity with two policy families across three heterogeneous robot embodiments. Policies trained solely on TacClaw-Infinity demonstrations transfer without embodiment-specific robot demonstrations, and a controlled comparison shows that tactile observations improve performance on contact-rich tasks.

\begin{table*}[t]
\centering
\caption{
Comparison with representative large-scale, UMI-style, and tactile manipulation datasets. ``Distributed'' indicates collection across multiple physical locations or contributing institutions; temporal, semantic, and spatial annotations denote within-episode structured supervision.
}
\label{tab:dataset_comparison}
\resizebox{\textwidth}{!}{
\begin{tabular}{lccccccccc}
\toprule
\textbf{Dataset} &
\textbf{Collection} &
\textbf{Distributed} &
\textbf{Episodes} &
\textbf{Hours} &
\textbf{Tasks} &
\textbf{Tactile} &
\textbf{Temporal} &
\textbf{Semantic} &
\textbf{Spatial} \\
\midrule

Open X-Embodiment~\cite{openx2024} &
Real-Robot &
$\checkmark$ &
1M+ &
-- &
527 &
$\times$ &
$\times$ &
$\times$ &
$\times$ \\

DROID~\cite{khazatsky2024droid} &
Real-Robot &
$\checkmark$ &
76K &
350 &
86 &
$\times$ &
$\times$ &
$\times$ &
$\times$ \\

UMI~\cite{umi} &
UMI &
\checkmark &
1.4K &
-- &
4 &
$\times$ &
$\times$ &
$\times$ &
$\times$ \\

\midrule

ViTaMIn~\cite{liu2025vitamin} &
Tactile UMI &
$\times$ &
747 &
-- &
6 &
$\checkmark$ &
$\times$ &
$\times$ &
$\times$ \\

exUMI~\cite{xu2025exumi} &
Tactile UMI &
$\times$ &
1.7K &
-- &
9 &
$\checkmark$ &
$\times$ &
$\times$ &
$\times$ \\

Touch in the Wild~\cite{zhu2025touchwild} &
Tactile UMI &
$\checkmark$ &
2.7K+ &
-- &
43 &
$\checkmark$ &
$\times$ &
$\times$ &
$\times$ \\

FreeTacMan~\cite{wu2025freetacman} &
Tactile UMI &
$\times$ &
10K+ &
-- &
50 &
$\checkmark$ &
$\times$ &
$\times$ &
$\times$ \\

\midrule

\textbf{TacClaw-Infinity (Ours)} &
\textbf{Tactile UMI} &
$\checkmark$ &
\textbf{1M+} &
\textbf{10K+} &
\textbf{1K+} &
$\checkmark$ &
$\checkmark$ &
$\checkmark$ &
$\checkmark$ \\

\bottomrule
\end{tabular}
}
\vspace{-4mm}
\end{table*}

\section{Related Work}
\label{sec:related_work}
% \yichi{need more citations.}
% \yichi{fix the multi-citation of wu2025freetacman and wu2026freetacman. }
% \yichi{reference table 1 in relatedwork section.}
\subsection{Large-Scale Robot Demonstration Datasets}

Large-scale robot demonstration datasets such as RoboNet, RT-1, BridgeData V2, DROID, and Open X-Embodiment have shown that scaling task, environment, embodiment, and demonstration diversity can substantially improve policy generalization~\cite{dasari2019robonet,brohan2022rt1,walke2023bridgedatav2,khazatsky2024droid,openx2024}. These resources also support generalist vision-language-action policies through large-scale visual, language, and action supervision~\cite{brohan2023rt2,kim2025openvla,openx2024}. Multimodal datasets such as RH20T further incorporate force, audio, and other sensing modalities~\cite{fang2023rh20t}. However, existing large-scale datasets remain largely vision-centric or depend on robot-side collection, while synchronized high-resolution tactile data remain scarce. This limits their ability to capture contact-rich interaction states such as slip, insertion, and grasp stability. In contrast, TacClaw-Infinity scales synchronized vision--tactile--language--action demonstrations across diverse environments, providing dense contact information and structured supervision for contact-rich manipulation.

\subsection{UMI-Style and Tactile Data Collection}
Embodiment-free collection provides a scalable alternative to robot-side demonstrations. UMI enables in-the-wild data collection with a handheld gripper while retaining observation and action interfaces compatible with downstream robot deployment~\cite{umi}. Subsequent work improves the scalability of this paradigm: FastUMI simplifies hardware and pose tracking~\cite{zhaxizhuoma2025fastumi}, FastUMI-100K scales collection to more than 100{,}000 multimodal demonstrations~\cite{liu2025fastumi}, and large-scale studies further highlight the importance of object and environment diversity for generalization and enable foundation-model pretraining for open-vocabulary and zero-shot cross-embodiment manipulation~\cite{lin2025datascaling,liu2026rdt2}.

Another line of work focuses on improving the fidelity and deployability of UMI data. VISTA, HiFi-UMI, and UMI-Bench address visual-domain mismatch, trajectory and synchronization accuracy, kinematic feasibility, and standardized real-robot evaluation, making robot-free demonstrations more reliable for downstream policy learning~\cite{yang2026vista,simpleai2026hifiumi,jin2026umibench}. Meanwhile, tactile UMI systems extend the predominantly visuomotor interface with direct measurements of physical interaction. ViTaMIn, FreeTacMan, and Touch in the Wild introduce portable visuo-tactile collection~\cite{liu2025vitamin,wu2025freetacman,zhu2025touchwild}, while exUMI, ViTaMIn-B, and TacUMI further explore tactile predictive learning, bimanual collection, and multimodal manipulation~\cite{xu2025exumi,li2025vitaminb,cheng2026tacumi}. More recent systems broaden tactile UMI toward online feasibility and recovery collection, richer force and multimodal sensing, dexterous hands, and whole-body manipulation~\cite{wu2026tamen,luo2026omniumi,li2026aetherock,xu2025dexumi,xu2026realdexumi,jang2026wtumi}.

Beyond collection interfaces, tactile-conditioned policies and foundation models have investigated multimodal representation learning, language-conditioned action generation, and visuo-tactile world modeling~\cite{liu2025vtdexmanip,hao2025tla,zhang2025vtla,zheng2026omnivta}. Nevertheless, existing work largely focuses on individual interfaces, task-specific datasets, or policy-learning mechanisms. As summarized in Table~\ref{tab:dataset_comparison}, jointly scaling synchronized vision--tactile--language--action data across distributed environments while providing temporal, semantic, and spatial supervision remains underexplored. TacClaw-Infinity addresses this gap by combining large-scale multi-site tactile UMI collection with structured contact-centric annotations for both single-hand and bimanual manipulation.

\subsection{Structured Annotation for Long-Horizon Manipulation}
% \yichi{augmentation this part. }

Long-horizon demonstrations contain multiple interaction stages and subtask transitions, motivating temporal segmentation methods such as MS-TCN, MS-TCN++, ASFormer, and ASRF~\cite{abufarha2019mstcn,li2023mstcnpp,yi2021asformer,ishikawa2021asrf}, as well as language-conditioned and VLM-based demonstration parsing~\cite{raj2024tasksegments,wang2025vlmseerobotdo}. However, these approaches are often specialized to particular datasets or task families. TacUMI further demonstrates that multimodal signals can improve tactile event segmentation~\cite{cheng2026tacumi}. In contrast, TacClaw-Infinity adopts a task-agnostic, contact-centric annotation framework that jointly provides temporal stages, segment-level language descriptions, and spatial target annotations, converting heterogeneous demonstrations into structured supervision for large-scale VTLA learning.
\section{Dataset}
\label{method}

\subsection{Dataset Overview}

TacClaw-Infinity is a large-scale structured Vision-Tactile-Language-Action (VTLA) dataset designed for contact-rich manipulation. It combines robot-independent tactile demonstration collection with scalable data curation and structured annotation. The dataset contains over \textbf{10,000 hours}, \textbf{1 million episodes}, and \textbf{1 billion frames} spanning more than \textbf{1,000 task types}, as reported in TABLE~\ref{tab:dataset_comparison}. Each episode provides synchronized visual observations, fingertip tactile signals, end-effector trajectories, and language descriptions, together with temporal, semantic, and spatial annotations. The corpus includes both single-hand and bimanual manipulation and covers diverse objects, actions, environments, and physical interaction patterns. 
% \yichi{we can ref table 1 here.}
\subsection{TacClaw and Data Collection Protocol}
\label{sec:device}
TacClaw is a tactile-enabled UMI-style handheld interface used as the primary data-acquisition platform for TacClaw-Infinity. An exploded view of the device is shown in Fig.~\ref{fig:teaser}. The system integrates a wrist-mounted camera, a stereo camera, an inertial measurement unit (IMU), a gripper encoder, and a DM-Tac X visuotactile sensor, which captures three-axis tactile signals from the contact region at up to 120~Hz. Operators can directly perform manipulation demonstrations without requiring a target robot during data collection, while accurately recording end-effector motion, gripper state, visual observations, and tactile feedback.

Gripper mechanics determine how much of the contact reaches the operator's hand. In UMI~\cite{umi}, the trigger-activated push mechanism offsets the fingers from the pinch point; sliding finger-loop designs, such as the wearable parallel-hand exoskeleton~\cite{lu2023visualtactile}, introduce a similar offset. Since operators regulate force by 
feel, this attenuation limits how lightly a demonstration can be performed. TacClaw instead lets the fingers act directly on the jaws through a short, rigid path. In a blindfolded study across three interfaces (Sec.~\ref{sec:subject_experiment}), the mean first-perceived contact force 
is $11.6$~gf with TacClaw, versus $49.8$~gf and $204.0$~gf for the trigger-driven and finger-loop mechanisms respectively.

Each TacClaw records a monocular wrist RGB stream, three co-registered fingertip tactile modalities---deformation, shear, and depth---and its end-effector trajectory, all synchronized at 30 Hz; bimanual episodes combine two devices with an egocentric view, together with an episode-level language description.

Pose is estimated by visual-inertial odometry from a stereo module with a hardware-synchronized IMU. To relate the two handheld devices and the 
egocentric camera in a common frame, each TacClaw carries a fiducial marker on its rear face, visible to the ego camera during collection. Observing both markers lets us register the three local coordinate systems into a shared reference, which improves relative pose accuracy and is what makes the cross-view annotation in Sec.~\ref{sec:pipeline} possible.
% \yichi{wrong reference}

TacClaw-Infinity is collected across many sites. Exploiting the portability of TacClaw, 200 devices were deployed to more than 20 organizations and 50 physical locations across homes, laboratories, industrial workplaces, retail stores, and outdoor environments. Tasks are either drawn from a shared task catalogue, which yields repeated demonstrations of common skills, or defined by collectors and collaborators according to their own research and application interests, which contributes a long tail of less frequent manipulation behaviors. 

% Table [X] summarizes the distribution of hours and episodes across environment types.

% To balance collection scale, repeatability, and environmental diversity, we 
% adopt two complementary acquisition protocols. \textit{Protocol-driven 
% collection} is conducted in centralized facilities using predefined task 
% specifications, providing high-throughput and repeatable demonstrations. 
% \textit{Distributed collection} exploits the portability of TacClaw to 
% acquire demonstrations in homes, laboratories, industrial workplaces, business places, outdoor environments, and so on, including tasks contributed by academic and industrial collaborators according to their own research and 
% application interests, substantially broadening the task and scene 
% distribution beyond centralized collection.

% \weihao{indroduce how many devices are used, distributed, organized}

% \todo{Long-tail data analysis @fubo} % 加入外发式采集的长尾数据分析 @富博

\subsection{Data Curation and Annotation Pipeline}
\label{sec:pipeline}
\begin{figure}[t]
    \centering
    \includegraphics[width=\columnwidth]
    {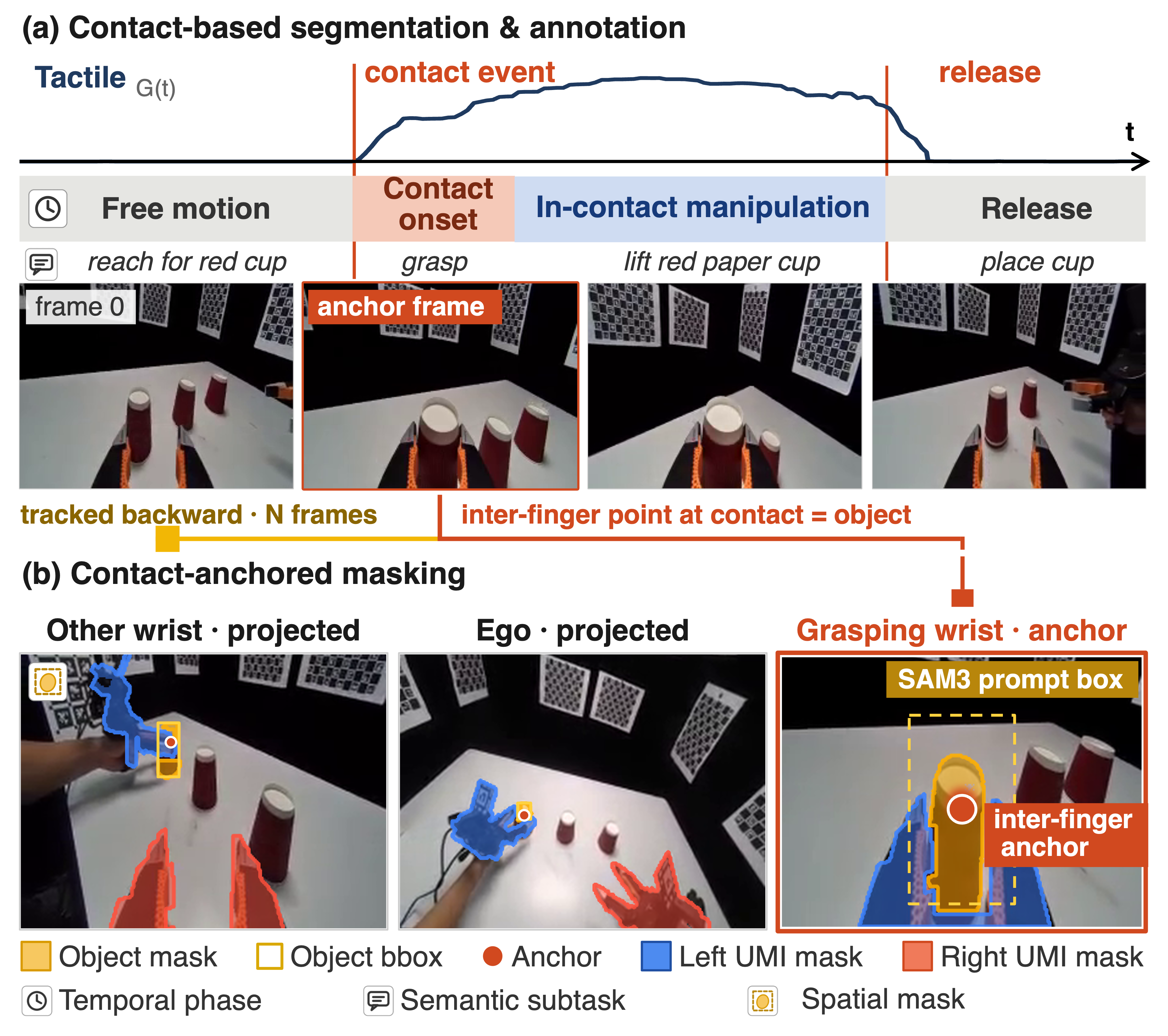}
    \caption{
    Contact-anchored structured annotation.
    (a) Tactile contact and release events partition a demonstration
    into manipulation stages and provide boundaries for segment-level
    language annotation.
    (b) The inter-finger point at contact provides a spatial anchor for identifying the manipulated object. Because the anchor is a single point in space, it projects into every camera of the rig, so each view is segmented independently from the same physical prior. The URDF silhouettes of both devices, rendered per view, serve as negative regions.
    }
    \vspace{-4mm}
    \label{fig:contact_annotation}
\end{figure}

Large-scale distributed data collection introduces variations in trajectory quality, sensing reliability, and robot executability. We therefore organize the pipeline into two stages: trajectory curation and data annotation.
For trajectory curation, we evaluate each demonstration in the action space of the target embodiment and filter by workspace reachability and manipulability.
For data annotation, as shown in Fig.~\ref{fig:contact_annotation}, tactile contact events serve as shared physical anchors for temporal, semantic, and spatial annotations.

\paragraph{Action-space Analysis}
To filter demonstrations that are not reliably executable by target robots, we develop a unified cross-platform action-space evaluation framework for Tianji Marvin M6S Lite, DM-Sparky One, and RealMan RM75-b, considering both reachability and manipulability. We first transform the demonstrated TCP poses into the base frame of each target robot for robot-specific kinematic evaluation.

\textbf{Reachability.}
Reachability measures whether a target robot can reproduce a demonstrated pose within its joint limits. For each target pose, we solve inverse kinematics to obtain the joint configuration $\mathbf{q}_t^*$ and reconstruct the corresponding pose through forward kinematics,
\begin{equation}
\hat{\mathbf{T}}_t=\operatorname{FK}(\mathbf{q}_t^*).
\end{equation}
A frame is considered reachable when the positional and rotational reconstruction errors satisfy $e_{p,t}<5~\mathrm{mm}$ and $e_{R,t}<3^\circ$, respectively. For bimanual trajectories, both arms must satisfy these criteria simultaneously. We report reachability as the proportion of reachable frames.

\textbf{Manipulability.}
For each reachable configuration, we further evaluate its local Cartesian motion capability using the isotropy of the weighted geometric Jacobian $\tilde{\mathbf{J}}_t$, where rotational components are scaled by $0.30~\mathrm{m\,rad^{-1}}$:
\begin{equation}
\eta_t=
\frac{\sigma_{\min}(\tilde{\mathbf{J}}_t)}
{\sigma_{\max}(\tilde{\mathbf{J}}_t)}.
\end{equation}
Higher $\eta_t$ indicates better-conditioned local motion. Based on perturbation experiments across all evaluated robots, we use $\eta_t=0.04$ as the low-manipulability threshold. We therefore define the low-manipulability rate for each arm as
\begin{equation}
\mathrm{Low\ Manip.}_{a}
=
\frac{
\sum_{t \in \mathcal{R}_{a}}
\mathbb{I}\!\left[\eta_{a,t}<0.04\right]
}{
|\mathcal{R}_{a}|
}
\times 100\%,
\end{equation}
where $a\in\{L,R\}$ and $\mathcal{R}_{a}$ denotes the set of reachable frames for the corresponding arm.

During subsequent policy training, trajectories identified as unreachable or low-manipulability are filtered out. We report the reachability and manipulability statistics of different robot embodiments in Sec.~\ref{action_validation}.

\begin{figure*}[]
    \centering
    \includegraphics[width=\textwidth]{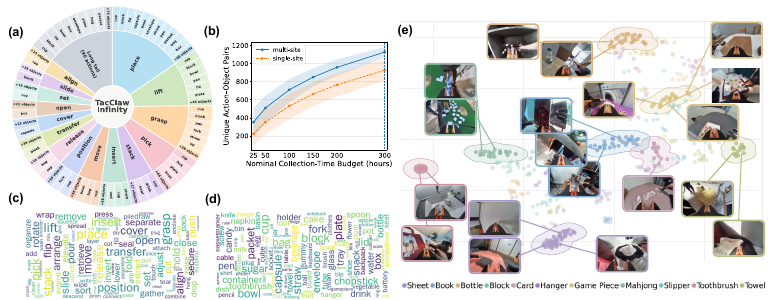}
    \caption{
    Statistics of TacClaw-Infinity.
    (a) Hierarchical distribution of manipulation actions and their associated object categories.
    (b) Unique action--object pairs under a matched collection-time budget, multi-site versus single-site; shaded bands are $95\%$ resampling intervals over 2{,}000 draws. 
    (c, d) Action and object vocabularies, with word size proportional to frequency.  
    (e) t-SNE of tactile features colored by object category.
    }
    \vspace{-4mm}
    \label{fig:dataset_statistics}
\end{figure*}

\paragraph{Contact-aware Temporal Segmentation}
As illustrated in Fig.~\ref{fig:contact_annotation}(a), we divide each demonstration into four interaction phases: free motion, contact onset, in-contact manipulation, and release.

To automatically identify these phases, we employ a temporal model to learn frame-wise phase transitions. Specifically, we formulate the two fingertip depth maps of each arm into a compact 8-D descriptor to provide a stable and contact-relevant representation for phase recognition, and feed it together with the two-view visual features and proprioceptive states into an ASFormer-based temporal model to predict the interaction phase of each frame. The model also predicts boundary scores, which are used to suppress weak transitions and merge overly short segments.

Finally, Qwen3-VL-Plus generates a subtask description for each segment based on its phase label and frames sampled from each camera.

\paragraph{Contact-anchored Spatial Annotation}
\label{sec:spatial_annotation}

Fig.~\ref{fig:contact_annotation}(b) illustrates the procedure. Each TacClaw is tracked as a rigid body, allowing us to render its URDF silhouette into every synchronized view and obtain per-view masks of both TacClaws.

For the manipulated object, we exploit a physical prior provided by tactile sensing: at the instant contact occurs, the object between the fingertips is directly identified as the manipulation target, without relying on semantic recognition. For each contact event, the URDF-defined inter-finger contact point provides a 3-D anchor, which is projected into every synchronized view to spatially localize the same target across cameras. In each view, we prompt SAM3 using the projected anchor, a local box of fixed physical extent, and an object noun generated by a VLM, while using the rendered TacClaw masks as negative regions. The resulting object mask is then propagated backward to the preceding contact-related event, providing temporally consistent target annotations.

\section{Dataset Statistics and Analysis}
We analyze TacClaw-Infinity in terms of scene and task diversity, action and object diversity, and tactile signal diversity to characterize both the breadth of the dataset and the richness of its physical interactions. We further quantify the benefit of distributed acquisition by comparing action--object coverage between single-site and multi-site collection under matched collection-time budgets. 
% \yichi{and we further analyse diversity differences between single-site and multi-site collection}

\paragraph{Scene and Task Diversity}
% 强调数量多，场景多，task多，并且室内外（有in the wild dat）且有/无SOP都有
TacClaw-Infinity spans more than \textbf{1,000 task types} collected at more than \textbf{50 distinct physical locations} across homes, university laboratories, industrial workplaces, retail stores, and outdoor environments, resulting in substantial variation in backgrounds, layouts, objects, operators, and manipulation conditions. The combination of predefined collection protocols and open-ended distributed acquisition provides both repeated demonstrations of common skills and a long tail of less frequent manipulation behaviors.
For example, the \textit{pick-cup} task is instantiated by 192 collectors across 8 organizations and 9 distinct scene settings, illustrating substantial diversity in collection contexts for the same manipulation skill.
% \yichi{If we have more detailed statistical data or more concreted examples to show our diversity, for example, for a task, e.g. wiping vase, we have xxx type background across xxx countries, organizations with over xxx collectors.}

\paragraph{Action and Object Diversity}
% task = action + object
As shown in Fig.~\ref{fig:dataset_statistics}(a), TacClaw-Infinity covers a broad hierarchical distribution of manipulation actions and associated object categories, with the full vocabularies shown in Fig.~\ref{fig:dataset_statistics}(c, d). Beyond grasping and placing, it includes contact-rich operations such as insertion, alignment, adjustment, wiping, rinsing, and plugging, many of which lie in the long tail of the action vocabulary (Fig.~\ref{fig:dataset_statistics}(c)). These action--object compositions span diverse geometric and physical interaction constraints, providing broad supervision for transferable manipulation learning.
For instance, the \textit{place-block} task spans 56 devices and 6 scenes, with blocks varying in 9 colors, 3 materials, and 5 shapes. It further covers 11 spatial relations and 11 behavior patterns, demonstrating rich variation in both object properties and manipulation behaviors within a single action--object composition.
% \yichi{If we have more detailed statistical data or more concreted examples to show our diversity, for example, for a task, e.g. wiping vase, we have and xxx type erasers, and xxx type vase.}

\paragraph{Tactile Signal Diversity}
% Beyond task-level diversity, TacClaw-Infinity exhibits substantial variation in tactile interaction patterns. We characterize this diversity using statistics such as contact ratio and contact area across different action primitives. Different actions exhibit distinct combinations of contact duration, spatial extent, and interaction intensity, ranging from sustained small-area grasping to brief but large-area contact events. This variation indicates that tactile sensing provides a rich and task-dependent learning signal rather than merely duplicating information available from vision.
Fig.~\ref{fig:dataset_statistics}(e) visualizes the tactile feature distribution for selected object categories. The substantial intra-category variation and partial inter-category separation reflect the diversity of contact states induced by different objects and manipulation behaviors.
% \yichi{mention that our data is collected globally. }

\paragraph{Action--Object Coverage of Multi-site Collection}
\label{sec:compositional}
% \yichi{more collector will lead to more randomness, so will lead to more action-object pair, this part do not show our mulit-site advantages. We can show that within the same task using the same time, multi-site can collect more diversity action trajectories, if possible. }
% 目的：外发式数采的数据分布优势说明，说明多个场所合作采集的action-object pair多样性比一个场所的大。
In this section, we compare the diversity achieved by collecting data across multiple sites with that achieved at a single site, under a fixed collection-time budget.
% 如何统计
To isolate the effect of site coverage from that of the task protocol, we restrict our analysis to open-ended recordings in which collectors selected their own tasks. 
For each budget ranging from 25 to 300~h, we randomly sample, without replacement, episodes whose total duration matches the given budget, drawing either from all sites (multi-site; i.e., the distributed collection described in Sec.~\ref{sec:device}, which spans multiple organizations and operators) or from a single site (single-site). We then count the number of distinct action--object pairs covered. This sampling procedure is repeated 2{,}000 times to estimate confidence intervals.

% 数据分析与推论，multi-site 采集的数据动作-物体对会更多
As reported in Fig.~\ref{fig:dataset_statistics}(b), multi-site collection consistently covers more action-object pairs than single-site collection, with $201$ more pairs ($95\%$ confidence interval: $[60, 335]$) at the budget of 300 hours, reflecting the greater diversity of scenes, operators, objects, and task preferences across sites.

% To separate the effect of collection breadth from that of collection volume, we resample sessions without replacement to a matched time budget (300~h, 2{,}000 repetitions) and compare sessions pooled across all sites (multi-site) against sessions from a single site (single-site). Coarse-grained coverage is comparable: the $95\%$ resampling intervals for canonical actions, manipulable objects, and the 14 task families all straddle zero. Compositional coverage is not: distributed collection observes $+200.9$ more action--object pairs ($95\%$ CI $[60.0, 335.0]$) and $+87.7$ more task-family--object pairs ($[30.0, 144.0]$), and the gap persists across budgets from 25 to 300~h. It also exposes more rare compositions in absolute terms ($809.2$ vs.\ $756.7$ pairs occurring at most five times). Under matched task families and matched sample size, scene-level DINOv2 embedding dispersion is higher as well ($\Delta=+1.06$, $[0.78, 1.35]$), although centralized collection yields more effective visual clusters under our clustering definition. Multi-site collection therefore contributes breadth of composition and context rather than a larger primitive vocabulary.

\begin{figure}[t]
    \centering
    \includegraphics[width=\linewidth]{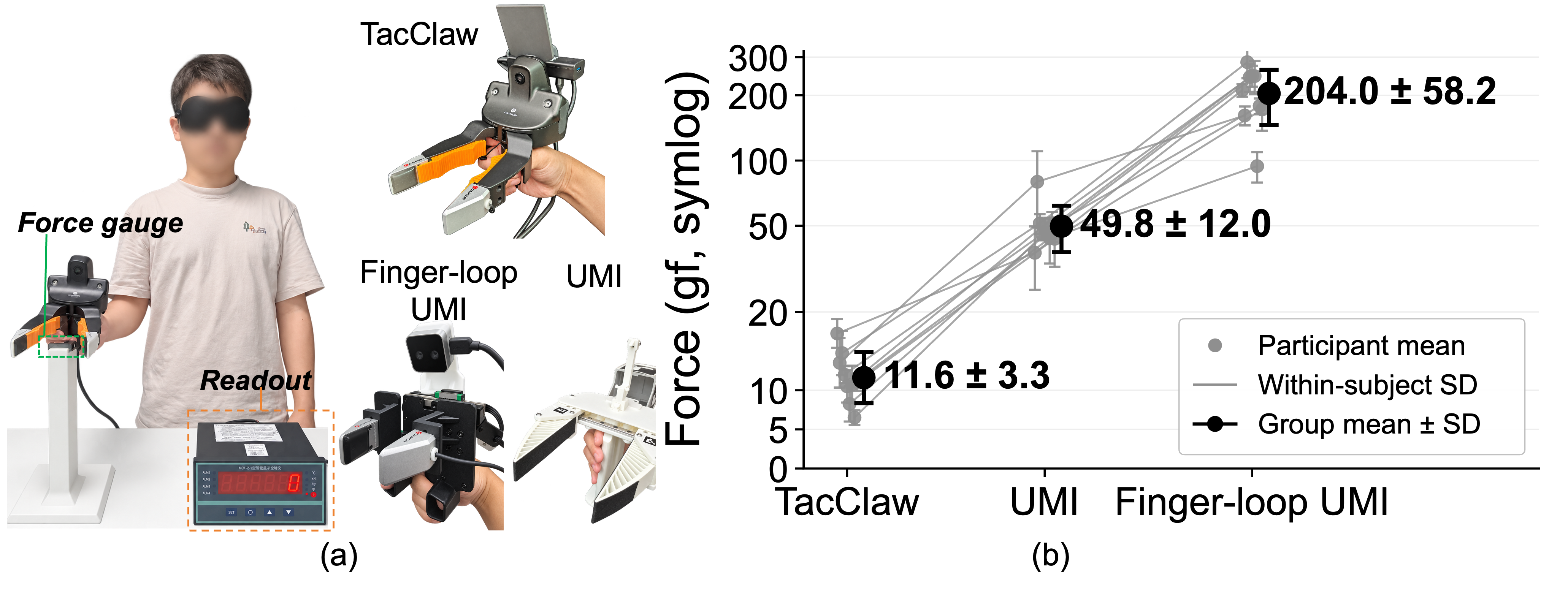}
    \caption{
    Haptic subject experiment ($n=9$).
    (a) Setup: a blindfolded participant closes the jaws onto a force gauge and says ``stop'' at first perceived contact; the three interfaces, which differ in how the fingers drive the jaws, are shown at right.
    (b) Perception threshold per interface. Gray markers denote participant means with within-subject standard deviation; black markers denote group mean $\pm$ standard deviation. The vertical axis uses a symlog scale 
    (linear below $20$~gf, logarithmic above). All interfaces were fabricated in-house; UMI follows the authors' released design files.
    }
    \label{fig:subject}
    \vspace{-5mm}
\end{figure}

\begin{figure*}[]
    \centering
    \includegraphics[width=\linewidth]{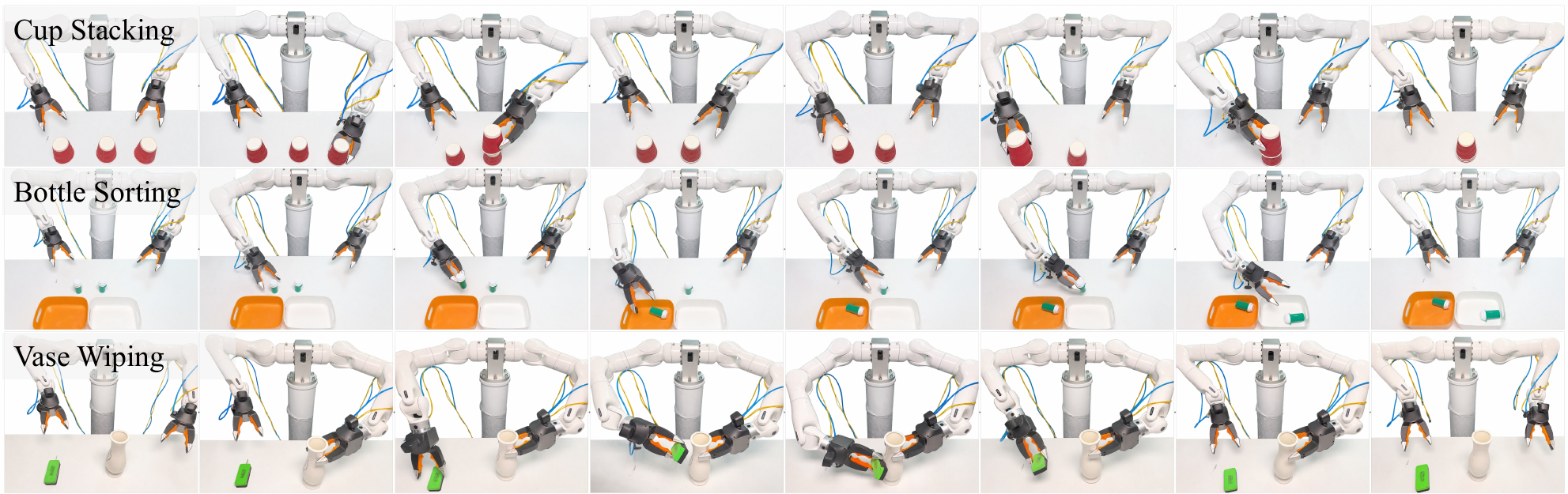}
    \caption{
    Representative execution sequences of the three bimanual long-horizon tasks: \textit{Cup Stacking}, \textit{Bottle Sorting}, and \textit{Vase Wiping}. The tasks respectively evaluate pick-and-place coordination, tactile-dependent state discrimination, and dynamic contact regulation.
    }
    \label{fig:task_overview}
\end{figure*}

\section{Experiments}
\label{sec:experiments}
% \yichi{need an overall here. }
We evaluate TacClaw-Infinity from multiple perspectives, including device-level haptic fidelity, action-space compatibility, annotation quality, and zero-robot cross-embodiment transfer. Together, these experiments assess both the reliability of our data collection and processing pipeline and the utility of the resulting dataset for downstream robot learning.

\subsection{Haptic Subject Experiment}          % 设备层，先讲
\label{sec:subject_experiment}
% 验证什么
To test the design argument of Sec.~\ref{sec:device}, we measure the lightest contact an operator can feel through each interface. 
% 怎么测试
Holding the interface with its jaws around the probe of a force gauge (Fig.~\ref{fig:subject}(a)), a blindfolded participant slowly closes the jaws and says ``stop'' at the first moment contact is felt;
the gauge reading at that moment is the \emph{perception threshold}.
We compare TacClaw with the original UMI gripper~\cite{umi} and a sliding-rail finger-loop gripper structurally similar to \cite{lu2023visualtactile}, all shown in Fig.~\ref{fig:subject}(a).
Nine participants (mean age $25.6$, range 20--33) each completed two practice trials and eight recorded trials per interface; the order of the three interfaces was randomized per participant.
% 测试结果

The perception threshold is $11.6 \pm 3.3$~gram-force (gf) with TacClaw, $49.8 \pm 12.0$~gf with UMI, and $204.0 \pm 58.2$~gf with the finger-loop gripper (Fig.~\ref{fig:subject}(b)).
All nine participants had the lowest threshold with TacClaw (Friedman $p<0.001$; pairwise Wilcoxon $p_{\mathrm{Holm}}=0.012$). 
% 结论分析
The finger-loop gripper is also the least consistent, which may be related to static friction in the rail. Of the three, only direct drive transmits contact to the hand without an intervening mechanism, and it yields the lowest threshold.

% Nine participants (age $25.6$ years, range $[20, 33]$), blindfolded,closed each of three handheld interfaces onto a force gauge and reported the force at which they first perceived contact between the fingertips(Fig.~\ref{fig:subject}). Besides TacClaw, we test the trigger-driven push mechanism of UMI~\cite{umi} and a sliding-rail finger-loop mechanism as used in N0-TacUMI~\cite{n0}. Participants received procedural training and two practice trials before the recorded session, and each condition was repeated $8$ times per participant in randomized order.

% The mean first-perceived contact force is $11.6 \pm 3.3$~gf with TacClaw, $49.8 \pm 12.0$~gf with UMI, and $204.0 \pm 58.2$~gf with the finger-loop variant. Every participant reported the lowest threshold with TacClaw($9/9$; Friedman test $\chi_F^2(2)=18.00$, $p<0.001$, Kendall's $W=1.00$, post-hoc exact two-sided Wilcoxon signed-rank
% tests with Holm correction, $p_{\mathrm{Holm}}=0.0117$ for all three pairwise comparisons). The finger-loop design is not only the least sensitive but also the most variable across participants, which is consistent with static
% friction in the sliding rail: the operator feels nothing until that friction is overcome. These results support the design argument in Sec.~\ref{sec:device}: the shorter and stiffer the path from the fingers to the jaws, the lighter the contact an operator can perceive and therefore regulate during demonstration.

\subsection{Action-space Validation}             % curation 有没有用 
\label{action_validation}
% Across all bimanual trajectory frames, DM-Sparky achieved complete action-space coverage when all whole-body degrees of freedom were enabled. Tianji achieved a bimanual reachability rate of 97.23\%, followed by RealMan at 95.15\%. All embodiments maintained low-manipulability rates below 5\%, indicating that most reachable poses remained sufficiently far from kinematic singularities.
As shown in Table~\ref{tab:cross_embodiment_action_space}, the collected demonstrations maintain high bimanual reachability across all three embodiments. DM-Sparky achieves 100\% coverage with full-body control, while Tianji and RealMan reach 97.23\% and 95.15\%, respectively. Meanwhile, low-manipulability rates remain below 5\% for all embodiments, indicating that most reachable poses stay away from poorly conditioned or near-singular configurations. Together, these results indicate that the collected demonstrations exhibit strong cross-embodiment kinematic compatibility and are largely executable across heterogeneous robot platforms.
% Together, these results support the effectiveness of our embodiment-aware action-space curation.
% \yichi{cannot support the efectiveness of our curation instead support that our sop is good.}

\subsection{Annotation Quality}   % 标注准不准
\label{sec:annotation_quality}
% \paragraph{Temporal annotation quality.}

% We manually annotated 141 demonstrations (41 single-arm and 100 bimanual), 
% yielding 173,995 per-arm frame labels. We evaluated the contact-phase segmenter 
% on 30 held-out sequences. Averaged over three seeds, the model achieves $63.9$ F1@25 
% and $54.3$ contact-centric F1@25, compared with $57.6$ and $47.7$, respectively, 
% for the same model without the tactile stream. Tactile evidence improves 
% contact-boundary recognition in particular, increasing the mean contact-onset recall from $58.6\%$ to $68.0\%$. 
% The main error for the release class is confusion with free motion, consistent with 
% the gradual decay of tactile responses.

\paragraph{Temporal}
% 使用人工标注数据作为真值测评含触觉的分割方法：
We evaluate the contact-phase segmenter of Sec.~\ref{sec:pipeline} on a manually annotated benchmark of 137 distinct demonstrations (41 single-arm, 96 bimanual), with 15 bimanual demonstrations designated for testing; each arm is evaluated as a separate sequence, yielding 30 test sequences.
% 测试对比结果，使用控制变量的方式横向对比了不同算法backbone 以及 同一our SOTA backbone带触觉和不带触觉的分割结果
Table~\ref{tab:contact_phase_segmentation} reports segmental F1 at 25\% overlap (F1@25), edit score, and frame accuracy, plus F1@25 restricted to the contact-onset and release classes (C-F1@25), the boundaries our annotation depends on. 

All baselines receive the same visual and proprioceptive input; MS-TCN++~\cite{li2023mstcnpp}, FACT~\cite{lu2024fact}, and LTContext~\cite{bahrami2023ltcontext} are standard segmentation architectures, and ASFormer (no tactile)~\cite{yi2021asformer} is our backbone without the tactile descriptor. 

The latter comparison isolates what tactile sensing con- tributes: +6.3 F1@25 and +6.6 C-F1@25, with contact-onset recall rising from 58.6\% to 68.0\%. Remaining errors concentrate on release, which is confused with free motion as tactile response decays gradually.

\paragraph{Spatial}
Coverage of the spatial annotation depends on how reliably the object can be tracked. On protocol-driven recordings, where the manipulated entity is a discrete graspable object, $98.5\%$ of frames inside the labelling window carry a mask. Coverage shows that a mask is produced, not that it is exact; mask quality is bounded by SAM3 tracking. Benchmarking mask accuracy and ablating its effect on downstream policies are left for future work.

% \yichi{how we calculate the 98.5. }

\paragraph{Semantic}
A $5\%$ random sample of the VLM-generated subtask descriptions was manually reviewed for correct identification of the action and object. Roughly $83\%$ passed; tasks containing failed samples are held back for manual re-annotation. Since review is sample-based, errors may still remain in other descriptions.

% \yichi{only discard less than 5\% is not effective, instead we can report the ratio of un-discard / all within the 5\%. }

\subsection{Zero-robot Cross-Embodiment Transfer}  % 主实验
\paragraph{Experiment Setup}
% embodiment and poicy class 
% 3 embodiment: Tianji, Realman, DM Sparky 
% 2 policy family: pi0.5, starVLA 
We evaluate the dataset and learning framework across three heterogeneous robot embodiments, including Tianji, RealMan, and DM-Sparky, to examine cross-embodiment transferability under different kinematic and hardware configurations. Two representative policy families including $\pi_{0.5}$ and StarVLA are considered. Using both policy families allows us to evaluate whether the collected data generalize across different VLA architectures rather than being specialized to a single policy design.

% tasks 双臂 长程 
% 1 Cup stacking 普通pick place 
% 2 Bottle sorting 静态触觉判断 
% 3 Vase wiping 动态触觉接触
We consider three representative bimanual long-horizon manipulation tasks with different levels of contact dependence, as shown in Fig.~\ref{fig:task_overview}. \textit{Cup stacking} mainly evaluates conventional pick-and-place capability and long-horizon coordination. \textit{Bottle sorting} requires the policy to infer object states from relatively static tactile observations during grasping and placement. \textit{Vase wiping} involves sustained and dynamically varying surface contact, requiring continuous tactile feedback for stable interaction throughout the wiping process. Together, these tasks cover free-space manipulation, static tactile perception, and dynamic contact regulation.

 % implemental details 
 % we train pi05 on 8 A100 and starVLA on 4 A100, 保持总batch size=64 
 % for all task, we train for 5 epochs, test 3/4/5 epcoh, and report the best ones 
 % for hyper-param, we follow the defaut settings of pi05 and starVLA
For implementation, $\pi_{0.5}$ is trained on 8 NVIDIA A100 GPUs, while StarVLA is trained on 4 NVIDIA A100 GPUs, with the total batch size fixed to 64 for both methods. All tasks are trained for five epochs, and checkpoints from epochs 3, 4, and 5 are evaluated under the same model-selection protocol, with the best-performing checkpoint reported. Unless otherwise specified, all hyperparameters follow the default configurations of $\pi_{0.5}$ and StarVLA. We use 1{,}600 training episodes for each task and conduct 50 evaluation trials per task. For tactile variants, tactile observations are introduced to both policy families using the same simple image-input formulation.

% \yichi{TODO: how many episodes we use?}
% 1600  for cup stacking
% 1600 for bottle sorting
% 1600 for vase wiping
% \yichi{TODO: how many times we test per task?}
% 50 trails for each task
% \yichi{TODO: how we add tactile to these two families. }
% tactile observations are introduced only through a simple image input formulation (as image)

\paragraph{Result analysis}
As shown in Table~\ref{tab:zero_robot_transfer}, policies trained solely on TacClaw demonstrations can be directly deployed across three heterogeneous robot embodiments without collecting embodiment-specific robot demonstrations, while still achieving strong manipulation performance. The same policy formulation transfers across Tianji Marvin M6S Lite, RealMan RM75-b, and DM-Sparky One despite substantial differences in kinematics and hardware configurations. This cross-embodiment generalization indicates that TacClaw trajectories preserve sufficiently accurate and robot-executable motion information, supporting their use for downstream robot learning.

The benefit of tactile sensing becomes particularly evident on contact-dependent tasks. Averaged across the three embodiments, adding tactile observations improves $\pi_{0.5}$ from 50.0\% to 74.0\% overall SR and StarVLA from 46.0\% to 72.7\%. The gains are concentrated on \textit{Bottle Sorting}, where SR increases by 44.7 and 43.3 percentage points for $\pi_{0.5}$ and StarVLA, respectively, and on \textit{Vase Wiping}, with corresponding gains of 22.0 and 38.7 points. Importantly, tactile observations are introduced only through a simple image-input formulation, without a specialized tactile architecture or controller. These results indicate that TacClaw provides sufficiently informative tactile signals to improve contact-state discrimination and sustained contact regulation across different policy families and robot embodiments.
% \yichi{TODO: Fill in the form with actual SR and analyze the SR here.}

\begin{table}[]
\centering
\caption{
Cross-embodiment reachability \& manipulability (\%).
}
\label{tab:cross_embodiment_action_space}
\setlength{\tabcolsep}{3pt}
\renewcommand{\arraystretch}{1.15}
\resizebox{\columnwidth}{!}{
\begin{tabular}{lcccc}
\toprule
\textbf{Embodiment} &
\makecell{\textbf{Left}\\\textbf{Reach.}} &
\makecell{\textbf{Right}\\\textbf{Reach.}} &
\makecell{\textbf{Bimanual}\\\textbf{Reach.}} &
\makecell{\textbf{Low Manip.}\\\textbf{L/R}} \\
\midrule

Tianji
& 99.20 & 98.02 & 97.23
& 2.5776 / 3.1611 \\

DM-Sparky (full-body)
& 100.0 & 100.0 & 100.0
& 0.0 / 0.0 \\

RealMan
& 97.38 & 96.79 & 95.15
& 4.3127 / 4.3526 \\

\bottomrule
\end{tabular}
}
\end{table}

\begin{table}[]
\centering
\caption{
Contact-phase segmentation on 30 held-out per-arm sequences. The metrics of
mean $\pm$ standard deviation over 3 seeds are presented (\%).
}
\label{tab:contact_phase_segmentation}

\setlength{\tabcolsep}{3pt}
\renewcommand{\arraystretch}{1.15}
\resizebox{\columnwidth}{!}{
\begin{tabular}{lcccc}
\toprule
\textbf{Method} &
\textbf{F1@25} &
\textbf{Edit} &
\textbf{Acc} &
\textbf{C-F1@25} \\
\midrule

MS-TCN++ &
$48.0\pm2.6$ &
$56.9\pm3.7$ &
$50.2\pm2.3$ &
$43.0\pm2.4$ \\

FACT &
$57.7\pm0.7$ &
$63.8\pm4.1$ &
$56.8\pm4.4$ &
$45.9\pm3.1$ \\

LTContext &
$55.4\pm4.7$ &
$63.3\pm4.8$ &
$62.7\pm4.9$ &
$45.4\pm4.1$ \\

\makecell[l]{ASFormer\\(no tactile)} &
$57.6\pm0.5$ &
$63.3\pm1.2$ &
$69.5\pm0.8$ &
$47.7\pm1.0$ \\

Ours &
$\mathbf{63.9\pm0.7}$ &
$\mathbf{67.3\pm2.0}$ &
$\mathbf{70.4\pm0.9}$ &
$\mathbf{54.3\pm0.6}$ \\

\bottomrule
\end{tabular}
}
\end{table}

\begin{table}[]
\centering
\caption{
Zero-robot cross-embodiment transfer results. All results are reported as success rates (SR, \%).
}
\label{tab:zero_robot_transfer}
\setlength{\tabcolsep}{3pt}
\renewcommand{\arraystretch}{1.15}
\resizebox{\columnwidth}{!}{
\begin{tabular}{llcccc}
\toprule
\textbf{Policy} &
\textbf{Embodiment} &
\makecell{\textbf{Cup}\\\textbf{Stacking}} &
\makecell{\textbf{Bottle}\\\textbf{Sorting}} &
\makecell{\textbf{Vase}\\\textbf{Wiping}} &
\textbf{Avg.} \\
\midrule

\multirow{3}{*}{$\pi_{0.5}$}
& Tianji    & 90 & 44 & 28 & 54.0 \\
& RealMan   & 86 & 42 & 16 & 48.0 \\
& DM Sparky & 82 & 44 & 18 & 48.0 \\

\midrule

\multirow{3}{*}{$\pi_{0.5}$ + Tactile}
& Tianji    & 94  & 96 & 54 & 81.3 \\
& RealMan   & 80  & 76 & 38 & 64.7 \\
& DM Sparky & 100 & 92 & 36 & 76.0 \\

\midrule

\multirow{3}{*}{StarVLA}
& Tianji    & 78 & 46 & 12 & 45.3 \\
& RealMan   & 78 & 38 & 8  & 41.3 \\
& DM Sparky & 80 & 54 & 20 & 51.3 \\

\midrule

\multirow{3}{*}{StarVLA + Tactile}
& Tianji    & 70 & 94 & 60 & 74.7 \\
& RealMan   & 76 & 84 & 58 & 72.7 \\
& DM Sparky & 84 & 90 & 38 & 70.7 \\

\bottomrule
\end{tabular}
}
\end{table}

\section{Conclusion}
\label{sec:conclusion}

% \fucai{ yichi, now it's ur time }

We presented \textbf{TacClaw-Infinity}, a large-scale structured Vision-Tactile-Language-Action dataset for contact-rich manipulation. By combining robot-independent tactile demonstration collection with distributed acquisition, scalable data curation, and structured annotation, TacClaw-Infinity brings together large-scale multimodal interaction data with temporal, semantic, and spatial supervision. Our evaluations validate the dataset from both data and model perspectives: the TacClaw interface provides sensitive contact feedback, the trajectories are kinematically compatible across embodiments, and the annotations provide contact-aware supervision. Policies trained on the resulting data further demonstrate cross-embodiment transfer across heterogeneous robots without requiring embodiment-specific demonstrations, while controlled comparisons highlight the utility of tactile observations for contact-rich manipulation. We hope TacClaw-Infinity provides a foundation for scalable visuotactile robot learning and encourages further research on learning physical interaction from diverse human demonstrations.  

% \clearpage
\bibliographystyle{IEEEtran}
\bibliography{refs}

@STRING{CORL    = {{CoRL}}}

@STRING{RSS     = {{RSS}}}

@STRING{ICRA    = {{ICRA}}}

@STRING{ICLR    = {{ICLR}}}

@STRING{NIPS    = {{NeurIPS}}}

@STRING{CVPR    = {{CVPR}}}

@STRING{IROS    = {{IROS}}}

@STRING{BMVC    = {{BMVC}}}

@STRING{WACV    = {{WACV}}}

@STRING{RAL     = {{IEEE RA-L}}}

@STRING{ICCV    = {{ICCV}}}

@inproceedings{dasari2019robonet,
  title={RoboNet: Large-Scale Multi-Robot Learning},
  author={Dasari, Sudeep and Ebert, Frederik and Tian, Stephen and Nair, Suraj and Bucher, Bernadette and Schmeckpeper, Karl and Singh, Siddharth and Levine, Sergey and Finn, Chelsea},
  booktitle=CORL,
  year={2020},
}

@inproceedings{brohan2022rt1,
  title={RT-1: Robotics Transformer for Real-World Control at Scale},
  author={Brohan, Anthony and Brown, Noah and Carbajal, Justice and Chebotar, Yevgen and Dabis, Joseph and Finn, Chelsea and Gopalakrishnan, Keerthana and Hausman, Karol and Herzog, Alex and Hsu, Jasmine and others},
  booktitle=RSS,
  year={2023},
}

@inproceedings{walke2023bridgedatav2,
  title={BridgeData V2: A Dataset for Robot Learning at Scale},
  author={Walke, Homer Rich and Black, Kevin and Lee, Abraham and Kim, Moo Jin and Du, Max and Zheng, Chongyi and Zhao, Tony and Hansen-Estruch, Philippe and Vuong, Quan and He, Andre Wang and Myers, Vivek and Fang, Kuan and Finn, Chelsea and Levine, Sergey},
  booktitle=CORL,
  year={2023},
}

@inproceedings{fang2023rh20t,
  title={RH20T: A Comprehensive Robotic Dataset for Learning Diverse Skills in One-Shot},
  author={Fang, Hao-Shu and Fang, Hongjie and Tang, Zhenyu and Liu, Jirong and Wang, Chenxi and Wang, Junbo and Zhu, Haoyi and Lu, Cewu},
  booktitle=ICRA,
  year={2024},
}

@inproceedings{openx2024,
  title={{Open X-Embodiment}: Robotic Learning Datasets and {RT-X} Models},
  author={{Open X-Embodiment Collaboration}},
  booktitle=ICRA,
  year={2024},
}

@inproceedings{brohan2023rt2,
  title={{RT-2}: Vision-Language-Action Models Transfer Web Knowledge to Robotic Control},
  author={Brohan, Anthony and others},
  booktitle=CORL,
  year={2023},
}

@inproceedings{kim2025openvla,
  title={{OpenVLA}: An Open-Source Vision-Language-Action Model},
  author={Kim, Moo Jin and Pertsch, Karl and Karamcheti, Siddharth and Xiao, Ted and Balakrishna, Ashwin and Nair, Suraj and Rafailov, Rafael and Foster, Ethan P. and Sanketi, Pannag R. and Vuong, Quan and Kollar, Thomas and Burchfiel, Benjamin and Tedrake, Russ and Sadigh, Dorsa and Levine, Sergey and Liang, Percy and Finn, Chelsea},
  booktitle=CORL,
  year={2025},
}

@article{liu2025fastumi,
  title={{FastUMI-100K}: Advancing Data-driven Robotic Manipulation with a Large-scale {UMI}-style Dataset},
  author={Liu, Kehui and Jia, Zhongjie and Li, Yang and Zhaxizhuoma and Chen, Pengan and Liu, Song and Liu, Xin and Zhang, Pingrui and Song, Haoming and Ye, Xinyi and Cao, Nieqing and Wang, Zhigang and Zeng, Jia and Wang, Dong and Ding, Yan and Zhao, Bin and Li, Xuelong},
  journal={arXiv preprint arXiv:2510.08022},
  year={2025},
}

@inproceedings{xu2025exumi,
  title={ex{UMI}: Extensible Robot Teaching System with Action-aware Task-agnostic Tactile Representation},
  author={Xu, Yue and Wei, Litao and An, Pengyu and Zhang, Qingyu and Li, Yong-Lu},
  booktitle=CORL,
  year={2025},
}

@inproceedings{liu2025vtdexmanip,
  title={VTDexManip: A Dataset and Benchmark for Visual-Tactile Pretraining and Dexterous Manipulation with Reinforcement Learning},
  author={Liu, Qingtao and Cui, Yufei and Sun, Zhenyu and Li, Guodong and Chen, Jianyu and Ye, Qiang},
  booktitle=ICLR,
  year={2025},
}

@inproceedings{khazatsky2024droid,
  title={DROID: A Large-Scale In-The-Wild Robot Manipulation Dataset},
  author={Khazatsky, Alexander and others},
  booktitle=RSS,
  year={2024},
}

@inproceedings{umi,
  title={Universal Manipulation Interface: In-The-Wild Robot Teaching Without In-The-Wild Robots},
  author={Chi, Cheng and Xu, Zhenjia and Pan, Chuer and Cousineau, Eric and Burchfiel, Benjamin and Feng, Siyuan and Tedrake, Russ and Song, Shuran},
  booktitle=RSS,
  year={2024},
}

@article{lu2023visualtactile,
  title={Visual-Tactile Robot Grasping Based on Human Skill Learning from Demonstrations Using a Wearable Parallel Hand Exoskeleton},
  author={Lu, Zhenyu and Chen, Lu and Dai, Hengtai and Li, Haoran and Zhao, Zhou and Zheng, Bofang and Lepora, Nathan F. and Yang, Chenguang},
  journal=RAL,
  volume={8},
  number={9},
  pages={5384--5391},
  year={2023},
}

@article{liu2025vitamin,
  title={ViTaMIn: Learning Contact-Rich Tasks Through Robot-Free Visuo-Tactile Manipulation Interface},
  author={Liu, Fangchen and Li, Chuanyu and Qin, Yihua and Shaw, Ankit and Xu, Jing and Abbeel, Pieter and Chen, Rui},
  journal={arXiv preprint arXiv:2504.06156},
  year={2025},
}

@article{wu2025freetacman,
  title={{FreeTacMan}: Robot-Free Visuo-Tactile Data Collection
    System for Contact-Rich Manipulation},
  author={Wu, Longyan and Yu, Checheng and Ren, Jieji and Chen, Li
    and Jiang, Yufei and Huang, Ran and Gu, Guoying and Li, Hongyang},
  journal={arXiv preprint arXiv:2506.01941},
  year={2025},
}

@inproceedings{zhu2025touchwild,
  title={Touch in the Wild: Learning Fine-Grained Manipulation with a Portable Visuo-Tactile Gripper},
  author={Zhu, Xinyue and Huang, Binghao and Li, Yunzhu},
  booktitle=NIPS,
  year={2025},
}

@inproceedings{cheng2026tacumi,
  title={TacUMI: A Multi-Modal Universal Manipulation Interface for Contact-Rich Tasks},
  author={Cheng, Tailai and Chen, Kejia and Chen, Lingyun and Zhang, Liding and Zhang, Yue and Ling, Yao and Hamad, Mahdi and Bing, Zhenshan and Wu, Fan and Sharma, Karan and Knoll, Alois},
  booktitle=ICRA,
  year={2026},
}

@article{li2025vitaminb,
  title={ViTaMIn-B: A Reliable and Efficient Visuo-Tactile Bimanual Manipulation Interface},
  author={Li, Chuanyu and Liu, Chaoyi and Wang, Daotan and Zhang, Shuyu and Li, Lusong and Zeng, Zecui and Liu, Fangchen and Xu, Jing and Chen, Rui},
  journal={arXiv preprint arXiv:2511.05858},
  year={2025},
}

@article{hao2025tla,
  title={TLA: Tactile-Language-Action Model for Contact-Rich Manipulation},
  author={Hao, Peng and Zhang, Chaofan and Li, Dingzhe and Cao, Xiaoge and Hao, Xiaoshuai and Cui, Shaowei and Wang, Shuo},
  journal={arXiv preprint arXiv:2503.08548},
  year={2025},
}

@article{zhang2025vtla,
  title={VTLA: Vision-Tactile-Language-Action Model with Preference Learning for Insertion Manipulation},
  author={Zhang, Chaofan and Hao, Peng and Cao, Xiaoge and Hao, Xiaoshuai and Cui, Shaowei and Wang, Shuo},
  journal={arXiv preprint arXiv:2505.09577},
  year={2025},
}

@article{zheng2026omnivta,
  title={OmniVTA: Visuo-Tactile World Modeling for Contact-Rich Robotic Manipulation},
  author={Zheng, Yuhang and Gu, Songen and Li, Weize and Zheng, Yupeng and Zang, Yujie and Tian, Shuai and Li, Xiang and Wu, Ruihai and Hao, Ce and Gao, Chen and others},
  journal={arXiv preprint arXiv:2603.19201},
  year={2026},
}

@inproceedings{raj2024tasksegments,
  title={Learning Temporally Composable Task Segmentations with Language},
  author={Raj, Divyanshu and Patil, Omkar and Gu, Weiwei and Baral, Chitta and Gopalan, Nakul},
  booktitle=IROS,
  year={2024},
}

@inproceedings{wang2025vlmseerobotdo,
  title={{VLM See, Robot Do}: Human Demo Video to Robot Action Plan via Vision Language Model},
  author={Wang, Beichen and Zhang, Juexiao and Dong, Shuwen and Fang, Irving and Feng, Chen},
  booktitle=IROS,
  year={2025},
}

@inproceedings{yi2021asformer,
  title={ASFormer: Transformer for Action Segmentation},
  author={Yi, Fangqiu and Wen, Hongyu and Jiang, Tingting},
  booktitle=BMVC,
  year={2021},
}

@inproceedings{abufarha2019mstcn,
  title={MS-TCN: Multi-Stage Temporal Convolutional Network for Action Segmentation},
  author={Abu Farha, Yazan and Gall, Juergen},
  booktitle=CVPR,
  year={2019},
}

@article{li2023mstcnpp,
  author  = {Shi-Jie Li and Yazan Abu Farha and Yun Liu and Ming-Ming Cheng and Juergen Gall},
  title   = {{MS-TCN++}: Multi-Stage Temporal Convolutional Network for Action Segmentation},
  journal = {IEEE Transactions on Pattern Analysis and Machine Intelligence},
  volume  = {45},
  number  = {6},
  pages   = {6647--6658},
  year    = {2023},
  doi     = {10.1109/TPAMI.2020.3021756}
}

@inproceedings{ishikawa2021asrf,
  title={Alleviating Over-Segmentation Errors by Detecting Action Boundaries},
  author={Ishikawa, Yuchi and Kasai, Seito and Aoki, Yoshimitsu and Kataoka, Hirokatsu},
  booktitle=WACV,
  year={2021},
}

@inproceedings{zhaxizhuoma2025fastumi,
  title={{FastUMI}: A Scalable and Hardware-Independent Universal Manipulation Interface with Dataset},
  author={Zhaxizhuoma and Liu, Kehui and Guan, Chuyue and Jia, Zhongjie and Wu, Ziniu and Liu, Xin and Wang, Tianyu and Liang, Shuai and Chen, Pengan and
    Zhang, Pingrui and Song, Haoming and Qu, Delin and Wang, Dong and Wang, Zhigang and Cao, Nieqing and Ding, Yan and Zhao, Bin and Li, Xuelong},
  booktitle=CORL,
  year={2025},
}

@inproceedings{lin2025datascaling,
  title={Data Scaling Laws in Imitation Learning for Robotic Manipulation},
  author={Lin, Fanqi and Hu, Yingdong and Sheng, Pingyue and Wen, Chuan and You, Jiacheng and Gao, Yang},
  booktitle=ICLR,
  year={2025},
}

@article{liu2026rdt2,
  title={{RDT2}: Exploring the Scaling Limit of {UMI} Data Towards Zero-Shot Cross-Embodiment Generalization},
  author={Liu, Songming and Li, Bangguo and Ma, Kai and Wu, Lingxuan and Tan, Hengkai and Ouyang, Xiao and Su, Hang and Zhu, Jun},
  journal={arXiv preprint arXiv:2602.03310},
  year={2026},
}

@article{yang2026vista,
  title={{VISTA}: Vision-Grounded and Physics-Validated Adaptation of {UMI} Data for {VLA} Training},
  author={Yang, Siyuan and Guo, Linzheng and Lu, Ouyang and Zhaxizhuoma and Zhang, Daoran and Wang, Xinmiao and Xiao, Ting and Yan, Fangzheng and Chen,
    Zhijun and Ding, Yan and Yu, Chao and Bai, Chenjia and Li, Xuelong},
  journal={arXiv preprint arXiv:2606.04708},
  year={2026},
}

@article{simpleai2026hifiumi,
  title={{HiFi-UMI}: Learning Deployable Manipulation Policies from High-Fidelity {UMI} Data Alone},
  author={{Simple AI} and Wei, Yuteng and Ma, Jinming and Wang, Jiawei and Zhou, Weitao and Zuo, Yushen and Rui, Ke and Li, Minglei and Zhang, Jinhao
    and Pan, Zhikang and Wang, Xiang and Jia, Haoran and Du, Huan and Zeng, Zicheng and Ma, Jun and Qin, Guiyu and Zhang, Di and Li, Xiaofei},
  journal={arXiv preprint arXiv:2607.25895},
  year={2026},
}

@article{jin2026umibench,
  title={{UMI-Bench 1.0}: An Open and Reproducible Real-World Benchmark for Tabletop Robotic Manipulation with {UMI} Data},
  author={Jin, Shi and Wang, Yuntian and Duan, Yuhui and Wu, Di and Dong, Gaoqi and Liu, Xiaohang and Li, Xiaotong and Jia, Hongfei and Zhang, Zehao
    and Wang, Tianyu and Jia, Zhongjie and Yao, Yuanqi and Bai, Chenjia and Zhaxizhuoma and Liu, Siao and Cao, Nieqing and Wang, Jin and Yu, Chao and Ding, Yan},
  journal={arXiv preprint arXiv:2606.10382},
  year={2026},
}

@article{wu2026tamen,
  title={{TAMEn}: Tactile-Aware Manipulation Engine for Closed-Loop Data Collection in Contact-Rich Tasks},
  author={Wu, Longyan and Ren, Jieji and Jiang, Chenghang and Zhou, Junxi and Peng, Shijia and Huang, Ran and Gu, Guoying and Chen, Li and Li,
    Hongyang},
  journal={arXiv preprint arXiv:2604.07335},
  year={2026},
}

@article{luo2026omniumi,
  title={{OmniUMI}: Towards Physically Grounded Robot Learning via Human-Aligned Multimodal Interaction},
  author={Luo, Shaqi and Li, Yuanyuan and Hu, Youhao and Yu, Chenhao and Xu, Chaoran and Zhang, Jiachen and Yao, Guocai and Huang, Tiejun and He, Ran
    and Wang, Zhongyuan},
  journal={arXiv preprint arXiv:2604.10647},
  year={2026},
}

@article{li2026aetherock,
  title={{AetheRock}: An Arm-Worn Robot Teaching System for Force-Guided Vision-Tactile Learning},
  author={Li, Hong and Xu, Yue and Tang, Yihan and Dong, Yankang and Liu, Chenyuan and Yu, Chenyang and Li, Xuyang and Huang, Siyuan and Shen, Yujun
    and Xue, Nan and Li, Yong-Lu},
  journal={arXiv preprint arXiv:2606.09777},
  year={2026},
}

@article{xu2025dexumi,
  title={{DexUMI}: Using Human Hand as the Universal Manipulation Interface for Dexterous Manipulation},
  author={Xu, Mengda and Zhang, Han and Hou, Yifan and Xu, Zhenjia and Fan, Linxi and Veloso, Manuela and Song, Shuran},
  journal={arXiv preprint arXiv:2505.21864},
  year={2025},
}

@article{xu2026realdexumi,
  title={{RealDexUMI}: A Wearable Universal Manipulation Interface for Dexterous Robot Learning},
  author={Xu, Chaoyi and Jiang, Yixuan and Huan, Jiahui and Fu, Yuhui and Zhou, Haoyu and Yuan, Weitian and Yu, Jiayi and Zhang, Wanpeng and Yuan,
    Haoqi and Lu, Zongqing},
  journal={arXiv preprint arXiv:2606.06033},
  year={2026},
}

@article{jang2026wtumi,
  title={{WT-UMI}: Tactile-Based Whole-Body Manipulation via Force-Supervised Contact-Aware Planning},
  author={Jang, Jaehwi and Gu, Zhaoyuan and Cueva, Alfred and Chai, Zimeng and Sheng, Junjie and Nguyen, Thong and Galundia, Himank and Wu, Yifan and
    Xue, Huishu and Legene, Isaac and Mediratta, Ojas and Doan, Davin and Collins, Andrew and Sadegh, Sarah and Kim, KyoungMok and Dhalbisoi, Rishita and Chen,
    Zun and Zhao, Ye},
  journal={arXiv preprint arXiv:2606.13232},
  year={2026},
}

@inproceedings{lu2024fact,
  title={{FACT}: Frame-Action Cross-Attention Temporal Modeling for Efficient Action Segmentation},
  author={Lu, Zijia and Elhamifar, Ehsan},
  booktitle=CVPR,
  year={2024},
}

@inproceedings{bahrami2023ltcontext,
  title={How Much Temporal Long-Term Context Is Needed for Action Segmentation?},
  author={Bahrami, Emad and Francesca, Gianpiero and Gall, Juergen},
  booktitle=ICCV,
  year={2023},
}
\end{document}